\documentclass[10pt,twocolumn,letterpaper]{article}

\usepackage{iccv}
\usepackage{times}
\usepackage{epsfig}
\usepackage{graphicx}
\usepackage{amsmath}
\usepackage{amssymb}
\usepackage[utf8]{inputenc} 
\usepackage[T1]{fontenc}    
\usepackage{url}            
\usepackage{booktabs}       
\usepackage{amsfonts}       
\usepackage{nicefrac}       
\usepackage{microtype}      
\usepackage{xcolor}         
\usepackage{amsfonts}
\usepackage{amssymb}
\usepackage{graphicx}
\usepackage{amsthm,amsmath}
\usepackage{mathrsfs}
\usepackage{indentfirst}
\usepackage{multirow}

\usepackage{cite}
\usepackage{algorithm}  
\usepackage{algorithmicx} 
\usepackage{algpseudocode}
\usepackage{tabularx} 
\usepackage{ragged2e} 
\usepackage{booktabs} 

\usepackage[accsupp]{axessibility} 
\usepackage{wrapfig}
\usepackage{caption}
\usepackage{subcaption}
\newcommand{\method}{SBCL\xspace}

\usepackage[pagebackref=true,breaklinks=true,letterpaper=true,colorlinks,bookmarks=false]{hyperref}
\title{Subclass-balancing Contrastive Learning \\for Long-tailed Recognition}

\iccvfinalcopy
\def\iccvPaperID{9664} 

\ificcvfinal\fi

\begin{document}

\title{Subclass-balancing Contrastive Learning for Long-tailed Recognition}

\author{%
Chengkai Hou$^{1}$, \
Jieyu Zhang$^{2}$, \
Haonan Wang$^{3}$,
Tianyi Zhou$^{4}$\thanks{Corresponding author} \\
$^1$Jilin university, 
$^2$University of Washington,
$^3$National University of Singapore\\
$^4$ University of Maryland\\
\texttt{ \small houck20@mails.jlu.edu.cn, jieyuz2@cs.washington.edu, haonan.wang@u.nus.edu}\\
\texttt{\small tianyi@umd.edu}
}
\maketitle
\ificcvfinal\thispagestyle{empty}\fi

\maketitle

\begin{abstract} 

Long-tailed recognition with imbalanced class distribution naturally emerges in practical machine learning applications. Existing methods such as data reweighing, resampling, and supervised contrastive learning enforce the class balance with a price of introducing imbalance between instances of head class and tail class, which may ignore the underlying rich semantic substructures of the former and exaggerate the biases in the latter. We overcome these drawbacks by a novel ``subclass-balancing contrastive learning (SBCL)'' approach that clusters each head class into multiple subclasses of similar sizes as the tail classes and enforce representations to capture the two-layer class hierarchy between the original classes and their subclasses. Since the clustering is conducted in the representation space and updated during the course of training, the subclass labels preserve the semantic substructures of head classes. Meanwhile, it does not overemphasize tail class samples, so each individual instance contribute to the representation learning equally. Hence, our method achieves both the instance- and subclass-balance, while the original class labels are also learned through contrastive learning among subclasses from different classes. We evaluate SBCL over a list of long-tailed benchmark datasets and it achieves the state-of-the-art performance. In addition, we present extensive analyses and ablation studies of SBCL to verify its advantages. Our code is available at \url{ https://github.com/JackHck/subclass-balancing-contrastive-learning}.
\end{abstract}
\vspace{-2.0em}
\section{Introduction}\label{sec:intro}

In reality, the datasets often follow the Zipfian distribution over classes with a long tail~\cite{zipf2013psycho,spain2007measuring}, $i.e.$, a few classes (head classes) containing significantly more instances than the remaining tail classes. 
Such tail classes could be of great importance for high-stake applications, \eg, patient class in medical diagnosis or accident class in autonomous driving ~\cite{cao2019learning,shen2015long}.
However, training on such class-imbalanced datasets can result in a severely biased model with
noticeable performance drop in classification tasks~\cite{wang2017learning,mahajan2018exploring,zhong2019unequal,ando2017deep,buda2018systematic,Collobert2008A,yang2019me}.

To overcome the challenges posed by long-tailed data, data resampling~\cite{ando2017deep,buda2018systematic,chawla2002smote,shen2016relay} and loss reweighing~\cite{byrd2019effect,cao2019learning,cui2019class,dong2018imbalanced} have been widely applied but they cannot fully leverage all the head-class samples. 
Very recent work discovered that 
supervised contrastive learning (SCL)~\cite{khosla2020supervised} can achieve state-of-the-art (SOTA) performance on benchmark datasets of long-tailed recognition~\cite{kang2020exploring,li2021targeted}. 
Specifically, the $k$-positive contrastive learning (KCL)~\cite{kang2020exploring} and its subsequent work targeted supervised contrastive learning (TSC)~\cite{li2021targeted} revamp SCL by encouraging the learned feature space to be class-balanced 
and uniformly distributed.
However, those methods enforcing class-balance often come with a price of instance-imbalance, \ie, each individual instance of tail classes would have much greater impact on model training than that of head classes.

Such instance-imbalance can result in significant degradation of the performance on long-tailed recognition for several reasons. 
On the one hand, the limited samples in each tail class might not be sufficiently representative of the whole class. So even a small bias of them can be enormously exaggerated by class-balancing methods and result in sub-optimal learning of classifiers or representations.
On the other hand, head classes usually have more complicated semantic substructures, e.g., multiple high-density regions of the data distribution, so simply downweighing samples of head classes and treating them equally can easily lose critical structural information. 
For example, images of a head class ``cat'' might be highly diverse in breeds and colors, which need to be captured by different features but downweighing or subsampling them may easily lose such information, while a tail class ``platypus'' might only contain a few similar images that are unlikely to cover all the representative features.
Therefore, it is non-trivial to enforce both class-balance and instance-balance simultaneously in the same method. 

Can we remove the negative impact of class-imbalance while still retain the advantages of instance-balance? 
In this paper, we achieve both through subclass-balancing contrastive learning (SBCL), a novel supervised contrastive learning defined on subclasses, which are the clusters within each head class, have comparable size as tail classes, and are adaptively updated during the training. 
Instead of sacrificing instance-balance for class-balance, our method achieves both instance- and subclass-balance by exploring the head-class structure in the learned representation space of the model-in-training. 
In particular, we propose a \emph{bi-granularity contrastive loss} that enforces a sample (1) to be closer to samples from the same subclass than all the other samples; and (2) to be closer to samples from a different subclass but the same class than samples from any other subclasses. While the former learns representations with balanced and compact subclasses, the latter preserves the class structure on subclass level by encouraging the same class's subslasses to be closer to each other than to any different class's subclasses. Hence, it can learn an accurate classifier distinguishing original classes while enjoy both the instance- and subclass-balance.  

In this paper, we apply \method for  several visual recognition tasks to demonstrate \method superiority over other previous works (\eg, KCL\cite{kang2020exploring}, TSC~\cite{li2021targeted}). 
To summarize, this paper makes the following contributions:
\begin{enumerate}
\item [\textbf{(a).}] We provide a new design principal of leveraging supervised contrastive learning for long-tailed recognition, \ie, aiming at achieving both instance- and subclass-balance instead of class-balance at the expense of instance-balance.

\item [\textbf{(b).}] We propose a novel instantiation of the aforementioned design principal, subclass-balancing contrastive learning (SBCL), which consists of two major components, namely, subclass-balancing adaptive clustering and bi-granularity contrastive loss.

\item [\textbf{(c).}]  Empirically, we  compare the \method against state-of-the-art methods on three visual tasks: image classification, object detection, and instance segmentation to demonstrate its effectiveness on handling class imbalance. We also conduct a series of experiments to analyze the efficacy of \method.

\end{enumerate}

\section{Background and notations}

\paragraph{Long-tailed recognition.} Long-tailed recognition aims to learn a classifier from a training dataset with long-tailed class distribution, \ie, a few classes contain many data (head
classes) while most classes contain only a few data
(tail classes), where the major challenge is to require model recognizing all classes equally well. Let $\mathcal{D} = \left\{x_i,y_i\right\}_{i\in [n]}$ be a long-tailed training
dataset, where $x_i$ denotes a sample and $y_i\in [C]$ denotes its label. 
Denote by $\mathcal{D}_k\subseteq\mathcal{D}$ the set of instances belonging to class $k$ and $n_k=|\mathcal{D}_k|$ is the number of samples in class $k$.
The total number of training samples over $C$ classes is $n=\sum_{k=1}^C n_k$. Without
loss of generality, we follow prior work ~\cite{kang2019decoupling,hong2021disentangling} to assume
that the classes are sorted by cardinality in decreasing order
($i.e.$, if $i < j$, then $n_{i} \geq n_{j}$), and $n_1 \gg n_C$. In addition, we define the imbalance ratio as $ \max_{k\in [C]}(n_k)/\min_{k\in [C]}(n_k) = n_1/n_C$.
Finally, let $f_{\theta}(\cdot)$ be a deep feature extractor, \eg, a neural network, parameterized by $\theta$ and $\mathbf{w}_c$ is the linear classifier of class $c$, then the classifier we aim to learn is $ h(x_i) =\mathop{\arg\max}_{c\in [C]} \mathbf{w}_c^{\top} f_{\theta}(x_i)$.


\paragraph{Supervised contrastive learning.}

Recent studies have shown that supervised contrastive learning (SCL)~\cite{khosla2020supervised} provides a strong performance gain for long-tailed recognition and its variants have achieved state-of-the-art (SOTA) performance~\cite{kang2020exploring,li2021targeted}.
Specifically, SCL learns the feature extractor $f_{\theta}(\cdot)$ via maximizing
the discriminativeness of positive instances, \ie, instances from the same class, and the learning objective for a single training data $(x_i, y_i)$ in a batch $\mathcal{B}=\{(x_i, y_i)\}_{i\in [N]}$, is
\begin{equation}\label{eq:scl}
\begin{small}
\begin{aligned}
 \mathcal{L}_{SCL} = \sum_{i=1}^N-\frac{1}{|\tilde{P_i} |} \sum_{z_p \in  \tilde{P_i}} \log \frac{\exp(z_i \cdot z_p^\top / \tau )}{\sum_{z_a \in \tilde{V_i} }\exp(z_i \cdot z_a^\top / \tau )},
\end{aligned}
\end{small}
\end{equation}
where $\tau$ is the temperature hyperparameter, $z_i=f_{\theta}(x_i)$ is the feature generated from $x_i$,
$V_i = \left\{z_i\right\}_{i\in[N]}\backslash \left\{z_i\right\}$ is the current batch of features except for $z_i$,
$P_i=\left\{z_j \in V_i : y_j=y_i\right\}$ is a set of instances with the same label as $x_i$.
Finally, let $\tilde{z_i}$ be the feature of $\tilde{x_i}$, the augmented version of $x_i$, and for any set $S_i$ indexed by $i$, we use  $\tilde{S}_i=S_i\cup\left\{ \tilde{z_i} \right\}$, \eg, $\tilde{V_i}=V_i\cup\left\{ \tilde{z_i} \right\}$.
However, for the long-tailed datasets, the feature spaces is dominated by head classes and thus have limited capability of semantic discrimination ~\cite{kang2020exploring}. To address this, the $k$-positive contrastive learning (KCL)~\cite{kang2020exploring} attempts to balance the feature space by keeping the number of positive instances in $\tilde{P_i}$ equal for each class, leading to the following loss
\begin{equation}\label{eq:kcl}
\begin{small}
\begin{aligned}
 \mathcal{L}_{KCL} = \sum_{i=1}^N-\frac{1}{k+1} \sum_{z_p \in  \tilde{P_{i}^{k}}} \log \frac{\exp(z_i \cdot z_p^\top / \tau )}{\sum_{z_a \in \tilde{V_i} }\exp(z_i \cdot z_a^\top / \tau )}
\end{aligned}
\end{small}
\end{equation}
where $P_{i}^{k}$ is a subset of $P_i$ with $k$ randomly drawn instances. 
Finally, the learned feature extractor $f_{\theta}(\cdot)$ is exploited in a sequel stage of training the classifier for long-tailed recognition~\cite{kang2020exploring,li2021targeted}.

\section{Methodology}
\begin{figure*}[t]
\centering
\includegraphics[width=0.97\textwidth]{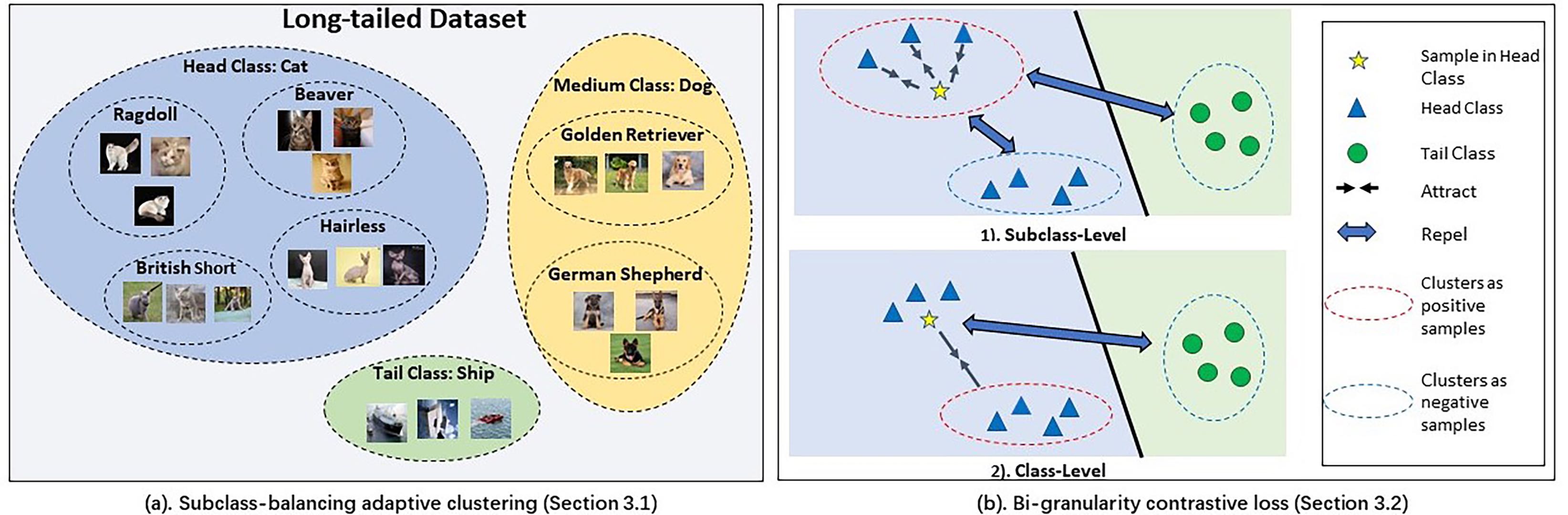}
\caption{Illustration of subclass-balancing 
contrastive learning (\method). It initially divides the head classes  into multiple subclasses of comparable size.
Then, during training, \method builds each sample to be closer to samples from the same subclass than samples from  different subclasses but the same class, which are also made to be closer than samples from different classes.}
\label{fig:method}
\end{figure*}

As mentioned above, KCL and its sequels~\cite{kang2020exploring,li2021targeted} balance the learning objective of SCL by picking the same number of positive instance for each class, \ie,  $|P_{i}^{k}|=k$ in Eq.~\ref{eq:kcl} no matter which class $x_i$ belongs to, however, we argue that such a class-balancing approach would inevitably introduce instance-imbalance: the instances of tail classes have much more chances to be engaged in the training than that of head class.
Specifically, assume each class has no less than $k$ instances, then the probability of an instance of class $c$ being selected as positive instance is $p(c) =\frac{k}{n_{c}}$; if the tail class $n_C$ has only $k$ instances and the imbalance ratio is $\frac{n_1}{n_C}=100$, then we have $p(1)=\frac{k}{100k}=0.01$ while $p(C)=1$.
We can see that when the instances of head class are selected once, that of tail class may already be trained 100 times.
Thus, the training is immensely biased towards the few samples in each tail class.
Besides, as tail classes only have very few instances that are not necessarily representative, the learned feature space might be unsatisfactory and sensitive to the training data of tail classes.

Here, we provide a new prospective of handling class-imbalance issue by contrastive learning: instead of aiming at class-balance at the expense of instance-imbalance, we propose to achieve both instance- and subclass-balance.
We argue that head classes typically contain more diverse instances and thus have richer semantics in the training dataset. Therefore, it might be wise to break down the head classes into multiple semantically coherent subclasses, each of which consists of similar number of instance as tail classes.
Built on this spirit, we develop subclass-balancing contrastive learning (SBCL), a new contrastive learning framework for long-tailed recognition (visualized in Figure~\ref{fig:method}) that achieves both instance- and subclass-balance. We present a concrete example that effectively demonstrates the superiority of \method compared to previous methods in Section~\ref{app:illustration}.




\subsection{Subclass-balancing adaptive clustering}
\label{sec:clustering}


We break down the class into several "subclasses" to attack the imbalanced phenomenon.
Particularly, given a class $c$ and the associated set of data $\mathcal{D}_c$, we employ a clustering algorithm of choice based on the features extracted by current feature extractor $f_{\theta}(\cdot)$ to divide $\mathcal{D}_c$ into $m_{c}$ subclasses/clusters. We use $\Gamma_{c}(x_i)$ to denote the cluster label of an instance $x_i$ of class $c$.
To ensure that the number of samples for each subclass is roughly the same,
we propose a new cluster algorithm to divide the unit-length feature vectors, \ie, the features output by $f_{\theta}(\cdot)$ with additional unit-length normalization.
The new proposed cluster algorithm is described in Algorithm \ref{alg:cluster}.

\begin{algorithm*}[t] 
\begin{small}
\caption{Subclass-balancing Adaptive Clustering}
  \label{alg:cluster}  
  \begin{algorithmic}[]  
   \Require 
      Sample set $\mathcal{S} =\left\{x_i\right\}_{i=1}^n$; 
      A threshold $M$; The number of iterations $K$;
   \Ensure  Cluster assignments for samples in $\mathcal{S}$
        \For{$k= 0$ to $K$}
        \If{ $k= 0$}
        \State Choose the cluster centers $y_j$ which are farther away from previously selected centers.
        \Else
        \State Update the cluster centers $y_j = \frac{1}{n_j} \sum_{i=1}^{n_j}x_i$.
        \Comment{$n_j$ is the  number of samples in a cluster}
        \EndIf
        \State Construct the cluster center set $\mathcal{C} =\left\{y_j\right\}_{j=1}^m$.
        \Comment{$m$ is the number of cluster centers}
        \While{$\mathcal{S} \neq \phi $}
        \Comment{Assign samples to centers $y_i$}
        \State Select the most similar pair $(x_i, y_j)= \mathop{\arg\max}\limits_{x \in \mathcal{S},y \in \mathcal{C}} \text{cosine-similarity}(x, y)$.
        \State Assign the sample $x_i$ to the center $y_j$.
        \State Delete the assigned sample $x_i$ from the sample set $\mathcal{S} = \mathcal{S} / \left\{x_i\right\}$.
        \If{ $ n_j \geq M$}
        \Comment{ Sample number in a cluster exceeds the threshold $M$}
          \State Delete the cluster center $y_j$ from the cluster center set $\mathcal{C} = \mathcal{C} / \left\{y_j\right\}$.
        \EndIf
        \EndWhile
        \EndFor
  \end{algorithmic}  
\end{small}
\end{algorithm*}  

In Algorithm~\ref{alg:cluster}, during the initialization of cluster centers, our approach gradually selects cluster centers by prioritizing the samples that are maximally distant from other existing cluster centers~\cite{arthur2007k}.
In the process of assigning vectors to their centers, we set a threshold $M$ as an upper limit of sample size in a cluster, which guarantees clusters of balanced sizes. 

Specifically, the threshold  $M$ is
\begin{equation}
\begin{footnotesize}
\begin{aligned}
M = \mathop{\max}(n_C , \delta)
\end{aligned}
\end{footnotesize}
\label{eq:cluster}
\end{equation}
where $n_C$ is the size of tail class and the hyperparameter $\delta$ controls the lower bound of sample size in clusters to prevent overly-small cluster.
Note that we only apply clustering algorithm to classes which contain multiple instances while the tail classes remain unchanged.
As a consequence, the size of each resultant cluster is similar to that of tail class $n_C$.
In addition, instead of only clustering once at the beginning, we update the cluster assignment adaptively based on the current feature extractor $f_{\theta}(\cdot)$ during training process and empirically show that such adaptive clustering outperforms only-once clustering in Section~\ref{sec:abl}.
We add an exhaustive description  of Algorithm~\ref{alg:cluster} and show the distribution of the sample size in clusters on the benchmark dataset in Appendix \ref{app: sample number}.

Then, by replacing the class labels of head classes used in SCL/KCL with the finer-grained cluster labels, we ensure the instance-balanced, \ie, each instance has similar chance of being selected regardless of its class.
By breaking down the head classes, which typically contain more diverse instances, into multiple semantically coherent subclasses, we achieve subclass-balanced (instead of class-balanced) while maintain the rich semantics rendered by head classes in training dataset.

\subsection{Bi-granularity contrastive loss} 
\label{sec:bi}

We now have two types of label for instances in head classes from different granularities: the coarse-grained \emph{class} label and the fine-grained \emph{cluster} label.
A direct consequence of replacing class label in SCL/KCL with cluster label is that we no longer distinguish instances from different head classes, and therefore the boundaries between classes might be blurry, leading to sub-optimal feature space.
As a  remedy, we combine the contrastive loss of both class label and cluster label into the following one and reuse the notations of Eq.~\ref{eq:scl}:

\begin{equation}\label{eq:loss}
\begin{small}
\begin{aligned}
  &\mathcal{L}_{\method}= - \sum_{i=1}^N \Big(\frac{1}{|\tilde{M}_i |} \sum_{z_p \in  \tilde{M_i}} \log \frac{\exp(z_i \cdot z_p^\top / \tau_1 )}{\sum_{z_a \in \tilde{V_i} }\exp(z_i \cdot z_a^\top / \tau_1 )}  \\
 & + \beta 
 \frac{1}{|\tilde{P_i}| - |M_i|} \sum_{z_p \in  \tilde{P_i}/M_i}\log \frac{\exp(z_i \cdot z_p^\top / \tau_2 )}{\sum_{z_a \in \tilde{V_i}/M_i }\exp(z_i \cdot z_a^\top / \tau_2 )} \Big)
\end{aligned}
\end{small}
\end{equation}
where $M_i = \left\{z_j \in P_i:  \Gamma_{y_i}(x_i)=\Gamma_{y_i}(x_j)\right\}$ is a set of instances with the same cluster label as $x_i$.
$\beta$ is a hyperparameter that balances these two loss terms.
The first term corresponds to the SCL loss with cluster labels, while the second term leverages the class label but does not consider the instances of the same cluster, \ie, the instances in $M_i$ are removed in the second term.
Such a design choice reflects the two types of positive instances for $z_i$: (1) the instance in the same cluster and (2) the instance of the same class but in different clusters.

According to previous studies~\cite{wang2021understanding, rince2022AAAI, PCL}, the temperature $\tau$ in contrastive loss is critical in controlling the local separation and global uniformity of the feature distribution.  
Specifically, for supervised contrastive
learning, a low temperature makes relative high penalty on feature distribution, that  actually encourages the features distribute more concentrically.
As the temperature increases, the relative penalty tends to be more uniform which uniformizes the distribution of the features.
Although the above objective explicitly considers the two types of label from different granularities, it still treats class and cluster label similarly.
Intuitively, we expect instances of the same subclass to form a more concentrated cluster in feature space than those of the same class, since subclass naturally indicates finer-grained semantic coherence.
To achieve this, we ensure the temperature $\tau_2>\tau_1$ and dynamically adjust $\tau_2$ for each class according to its current level of concentration of the instances' feature.
Following \cite{PCL}, for class $c$ we define $\phi(c)$ as
\begin{equation}\label{eq:t1}
\begin{small}
\begin{aligned}
\phi(c) = \frac{\sum_{i=1}^{n_c}\Vert z_i-t_{c}\Vert_2}{n_c \log(n_c + \alpha)}
\end{aligned}
\end{small}
\end{equation}
where $t_{c}$ is the centroid for the class $c$, $\alpha$ is a hyperparameter to ensure that $\phi(c)$ is not overly-large, and $z_i$ corresponds to instances of class $c$. 
From the formulation, we can see that if the current averaged distance to the class centroid is large 
or the class contains fewer data, thus the temperature  will be set large to adopt the feature distribution of class $c$ during the training process. 
Then we define the temperature of class $c$ as
\begin{equation}\label{eq:t2}
\begin{small}
\begin{aligned}
\tau_2(c) = \tau_1 \cdot \exp\left(\frac{\phi(c)}{\frac{1}{C}\sum_{i=1}^{C}\phi(i)}\right)
\end{aligned}
\end{small}
\end{equation}
such that $\tau_2(c)$ for class label is always larger than $\tau_1$ for cluster label (since $\phi(c)>0$) and could reflect the current level of concentration of the instances in a class.
In particular, the proposed $\tau_2(c)$ encourages the features of instances in class $c$ to form a less tight cluster than that of a subclass (by $\tau_2(c)>\tau_1$) while adaptively adjust the temperature to prevent an overly-loose/dense cluster.

\subsection{Training algorithm}\label{warm up}

Here, we describe the overall training process of subclass-balancing contrastive learning and the algorithm can be found in Algorithm~\ref{alg:Framwork}.
First, the adaptive clustering (Section~\ref{sec:clustering}) could be noisy at the early stage of training~\cite{PCL,wang2021unsupervised}. 
Thus, we warm-up the feature extractor $f_{\theta}(\cdot)$ by a few epochs of training on ordinary SCL or KCL loss.
In addition, our algorithm involves two adaptively-adjusting parts, namely, the cluster assignment and the temperature $\tau_2(c)$ for each head class $c$.
Instead of updating these every epoch, we use a hyperparameter $K$ as the update interval, \ie, we update the cluster assignment and the temperature based on the current learned $f_{\theta}(\cdot)$ every $K$ epoch.

\begin{algorithm}[http]
\begin{small}
  \caption{Training Algorithm}
  \label{alg:Framwork}  
  \begin{algorithmic}[1]
   \Require 
      Dataset $\mathcal{D} = \left\{x_i,y_i\right\}_{i\in [n]}$; ; The update interval of cluster assignment $K$;
      The number of warm-up epoch $T_0$; The total number of epoch $T$; The hyperparameters $\beta$ and $\delta$. 
   \Ensure
       A trained feature extractor $f_{\theta}(\cdot)$
    \State Initialize the model parameters $\theta$
    \State Train $f_{\theta}(\cdot)$ with SCL/KCL for $T_0$ epochs \Comment{Warm-up stage}
    \For{$t= T_0$ to $T$}
        \If{$t \% K == 0$} \Comment{Update cluster and termperature}
        \State Update the cluster assignment based on the current feature extractor $f_{\theta}(x)$
        \State Update the temperture $\tau_2$ for each head class using Eq.~\ref{eq:t1} and Eq.~\ref{eq:t2}
        \EndIf 
        \State Train $f_{\theta}(\cdot)$ using Eq.~\ref{eq:loss} \Comment{Subclass-balancing contrastive learning}
    \EndFor 
  \end{algorithmic}  
\end{small}
\end{algorithm}

\section{Experiment}
\label{sec:experiment}
\subsection{Experimental setup}\label{sec:expimental settings}

\paragraph{Datasets.} We consider three commonly used long-tailed recognition benchmark datasets: CIFAR-100-LT~\cite{cao2019learning}, ImageNet-LT~\cite{liu2019large}, and iNaturalist 2018~\cite{van2018inaturalist}.
The  CIFAR-100-LT and ImageNet-LT datasets are artificially generated long-tailed datasets from the class-balanced datasets~\cite{krizhevsky2009learning,russakovsky2015imagenet}, and the iNaturalist 2018 dataset is a large-scale real-world dataset that exhibits long-tailed imbalance.

\paragraph{Baselines.}
We consider baseline methods of the following three categories: 
(1) class-balancing classifiers, including $\tau$-norm,  LWS and cRT~\cite{kang2019decoupling}, which fixes the representation which trained by cross-entropy loss and trains the classifier with class-balanced sampling;
(2) one-stage balancing loss, including CB loss~\cite{cui2019class}, Focal loss~\cite{lin2017focal},  and LDAM loss~\cite{cao2019learning}. 
These supervised distribution-aware loss makes the model to pay more attention on the minority class during training.
(3) contrastive learning methods, including SCL~\cite{khosla2020supervised},  KCL~\cite{kang2020exploring}, SwAV~\cite{caron2020unsupervised}, PCL~\cite{PCL} and TSC~\cite{li2021targeted} which train a feature extractor with the contrastive loss and then learn a classifier given the trained feature extractor.
\vspace{-1.0em}

\paragraph{Evaluation protocol.} 
Following~\cite{kang2020exploring,kang2019decoupling,li2021targeted}, we implement \method, as well as other contrastive learning methods, in a two-stage framework. In the first stage, we train the feature extractor with a contrastive learning method, while in the second stage, we train a linear classifier on top of the learned representation. 
Specifically, for CIFAR-100-LT dataset, the linear classifier is trained with LDAM loss and class re-weighting~\cite{cao2019learning}. 
For ImageNet-LT and iNaturalist 2018 datasets, the linear classifier is trained with CE loss and class-balanced sampling~\cite{kang2019decoupling}. 
All results are averaged over 5 trials with different random seeds.
We mainly report the overall top-1 accuracy. 
For the two large datasets, ImageNet-LT and iNaturalist 2018 datasets, following the previous work~\cite{liu2019large}, we also report the accuracy of three disjoint subsets: Many-shot classes (classes with more than 100 samples), Medium-shot classes (classes with 20 to 100 samples), and Few-shot classes (classes under 20 samples).
We show the implementation details and important hyperparameters in the Appendix~\ref{apd:implementation_detail}. 

\subsection{Main results}

\begin{table*}[t]
\centering
\caption{Performance comparison on ImageNet-LT and iNaturalist 2018 datasets. Top-1 accuracy of ResNet-50 ~\cite{he2016deep} is reported. 
The "Many", "Medium" ,"Few" and "All" denotes different groups. $\dagger$ denotes our reproduced results of PCL, SwAV, and BYOL based on their official code. Other baselines' results on ImageNet-LT and iNaturalist 2018 are copied from  Li \etal~\cite{li2021targeted}.}
\label{table:large_dataset}
\begin{small}
\resizebox{0.7\linewidth}{!}{
\begin{tabular}{l|c|c|c|c|c|c|c|c}
\toprule[1pt]
Backbone & \multicolumn{4}{c|}{ImageNet-LT} & \multicolumn{4}{c}{iNaturalist 2018}  \\
\hline
Methods  & Many  &  Medium &  Few  &  All & Many  &  Medium &  Few  &  All    \\ 
\hline
CE  &64.0 &33.8 &5.8&41.6 &72.2  &63.0 &57.2&61.7 \\
{Focal loss~\cite{lin2017focal}}  &51.0 &40.8 &20.8 &43.7 & -  &- &- & 61.3 \\
{CB-Focal~\cite{cui2019class}}  &- &- &-&- & -  &- &- & 61.1 \\
{LDAM-DRW~\cite{cao2019learning}}  &60.4 &46.9 &30.7&49.8 &  -  &- &- & 64.6 \\
\hline
{OLTR~\cite{liu2019large}} &35.8 &32.3 &21.5 &32.2 &  59.0  &64.1 &64.9& 63.9 \\
{$\tau$-norm~\cite{kang2019decoupling}}   &  56.6   &44.2 &27.4&46.7  &  71.1 &68.9 &69.3 &69.3 \\
{cRT~\cite{kang2019decoupling}} &58.8 &44.0 &26.1 &47.3  &73.2  &68.8 & 66.1 & 68.2 \\
{LWS~\cite{kang2019decoupling}}   & 57.1 &45.2& 29.3 &47.7  & 71.0 &69.8 &68.8& 69.5 \\

\hline
{PCL$\dag$~\cite{PCL}} &34.7  &26.1  &12.3 &27.5 &48.5  & 45.9 &41.7 & 44.5 \\
{SwAV$\dag$~\cite{caron2020unsupervised}} &37.5 &28.3  &15.6 &30.1 &51.9  &48.4  &43.7 &47.0  \\
{BYOL$\dag$~\cite{grill2020bootstrap}} &37.7 &28.9  &16.3 &30.6 &52.3  &48.6  &44.1 &47.2 \\
{SCL~\cite{khosla2020supervised}} &61.4   & 47.0 &28.2 &49.8 &- &-& -&66.4  \\
{KCL~\cite{kang2020exploring}}   & 62.4 &49.0& 29.5 &51.5 &  -  &-&-& 68.6 \\
{TSC~\cite{li2021targeted}}   & 63.5   &49.7&30.4  &52.4 & 72.6   & 70.6& 67.8 &69.7 \\

\hline 
{\method} &\textbf{63.8} &\textbf{51.3} &\textbf{31.2} &\textbf{53.4}  &\textbf{73.3} &\textbf{71.9} &\textbf{68.6}&\textbf{70.8} \\
\bottomrule[1pt]
\end{tabular}}
\end{small}

\end{table*}

\begin{table}[t]
\centering
\caption{Performance comparison on CIFAR-100-LT. Top-1 accuracy of the ResNet-32~\cite{he2016deep} under different imbalance ratios is reported. We also report the accuracy of  our re-implemented important baselines ($\dag$) in same setting on CIFAR-100-LT.
The columns of "Statistic (IR 100)" are the results of different disjoint subsets on CIFAR-100-LT with imbalance ratio being 100.  }
\label{table:cifar}
\begin{small}
\resizebox{1.\linewidth}{!}{
\begin{tabular}{l|c|c|c|c|c|c}
\toprule[1pt]
\multirow{2}*{Method}&\multicolumn{3}{c|}{Imbalance Ratio}&\multicolumn{3}{c}{Statistic (IR 100)} \\
\cline{2-7}
 &100 &50 &10&Many &Medium &Few \\ 
\hline
CE     &38.3      &  43.9  & 55.7 &65.2 &37.1 &9.1 \\
CB-CE~\cite{cui2019class}   &38.6 &44.6& 57.1&- &-& - \\
Focal Loss~\cite{lin2017focal}    &38.4     &44.3  &55.8 &65.3 & 38.4 & 8.1     \\
CB-Focal~\cite{cui2019class}   &38.7 &45.2   &58.0 &65.0 & 37.6 &10.3    \\
CE-DRW~\cite{cao2019learning}  &41.4 &45.3 &58.1&- &- &-  \\
CE-DRS~\cite{cao2019learning}   &41.6 &45.5&58.1&- &-& -  \\
LDAM~\cite{cao2019learning}   &39.6 &45.0&56.9&- &-& - \\
LDAM-DRW~\cite{cao2019learning}   &42.0 &46.6 & 58.7&61.5 &41.7 &20.2\\
M2m-ERM~\cite{kim2020m2m}  &42.9 &- &58.2&- &-& -  \\
M2m-LDAM~\cite{kim2020m2m}&43.5 &- &57.6 &- &-& -  \\
\hline
{cRT~\cite{kang2019decoupling}} &43.3 &46.8 &58.1&64.0 &44.8 &18.1\\
{LWS~\cite{kang2019decoupling}}& 43.1 &46.4 &58.1&- &-& - \\
\hline
SCL~\cite{khosla2020supervised} $\dag$&42.1 &45.2 &54.8  &62.8&42.0&18.4\\
KCL~\cite{kang2020exploring}&42.8 &46.3 &57.6&-&-&- \\
KCL$\dag$&42.8&46.4 &57.5&63.4&42.5&19.2 \\
TSC~\cite{li2021targeted}&43.8 &47.4 &\textbf{59.0}&-&-&-\\
TSC$\dag$ &43.5 &47.6 &58.7&63.7&43.2&20.4\\
\hline
\method &\textbf{44.9}&\textbf{48.7}&57.9&\textbf{64.4}&\textbf{45.3}&\textbf{22.2}\\
\bottomrule[1pt]
\end{tabular}}
\end{small}
\vspace{-1.0em}
\end{table}

The results on ImageNet-LT and iNaturalist 2018 are in Table \ref{table:large_dataset}. 
We can see that \method outperforms the baselines with a large margin over the two datasets. 
In addition, on iNaturalist 2018 dataset, \method outperforms the previous SOTA method by 0.7\% on Many, 1.3\% on Medium, 0.8\% on Few and 1.1\% on All,  which shows the effectiveness of the proposed method in solving real-world long-tailed recognition problems such as natural species classification.
Besides, \method is also better than existing contrastive learning method like KCL and TSC for all class splits, which demonstrates the effectiveness of the design principal of pursuing both instances- and subclass-balance in contrastive learning.
Table \ref{table:cifar} summarizes the results on  CIFAR-100-LT dataset.
For CIFAR-100-LT dataset, \method outperforms previous SOTA methods except for imbalance ratio 10. 
We hypothesize that it is because the tail class of CIFAR-100-LT with imbalance ratio 10 has multiple samples, which makes it hard to distinguish the performance of methods on the long-tailed recognition.

\subsection{Performance on Other Visual Tasks}
\label{sec:object}

\begin{table}[h]
\centering
\begin{small}
\caption{Object detection results on PASCAL VOC.}
\label{table:voc}
\resizebox{1.0\columnwidth}{!}{\begin{tabular}{l|c|c|c|c|c|c}
\toprule[1pt]
\multirow{2}*{Method}&\multicolumn{3}{c|}{ImageNet}  &\multicolumn{3}{c}{ImageNet-LT} \\
\cline{2-7}
~ &AP$_{50}$  & AP &AP$_{75}$ &AP$_{50}$& AP &AP$_{75}$ \\ 
\hline
random init. &60.2 &33.8 &33.1 &60.2 &33.8 &33.1\\
 CE    &  81.3 &53.7 &59.2 &76.5 &48.5 &51.0  \\
 CL~\cite{he2020momentum} &81.3  &56.1  &62.7  &78.2 &51.5  &56.5\\
 KCL~\cite{kang2020exploring}& \textbf{82.3} &55.5 & 62.1  &79.7  &52.6  &57.9 \\
 \hline
\method &81.9 &\textbf{56.2}&\textbf{62.8}&\textbf{80.6} &\textbf{53.4}  &\textbf{58.8} \\
\bottomrule[1pt]
\end{tabular}}
\end{small}
\end{table}

There is a recent trend of using the contrastive learning to pretrain a feature extractor for downstream visual tasks other than image classification~\cite{he2020momentum}.
We are curious about two questions: (1) when the pretraining dataset is class-imbalanced, how the downstream performance is affected? (2) In such case, can our \method improve the learned feature extractor over existing contrastive learning baselines?
To answer these questions, we use the object detection task of PASCAL VOC dataset as the evaluation suite and use ImageNet/ImageNet-LT datasets as class-balanced/-imbalanced pretraining datasets.
Following~\cite{kang2020exploring,he2020momentum}, we first pretrain a feature extractor on ImageNet/ImageNet-LT then further finetune it for the downstream object detaction tasks using Faster R-CNN~\cite{ren2015faster} with R50-C4 backbone.

The experiment results are shown in Tables \ref{table:voc}. 
From the results, we can see that pretraining on class-balanced data (ImgeNet) leads to consistently better results than that on class-imbalanced dataset (ImageNet-LT)
pretraining the model on the  ImageNet and ImageNet-LT datasets by the \method can perform slightly better than other baselines. 
In addition, the proposed \method significantly outperforms baselines on class-imbalanced pretraining dataset, while achieve comparable performance on class-balanced ones.
For the representation which trained on the full ImageNet dataset, the performance advantage is not obvious.
In Appendix~\ref{appendix:coco}, we show additional experimental results of object detection and instance segmentation  on COCO~\cite{lin2014microsoft} dataset  and \method also outperforms other baseline methods. 
Thus, we conclude that the proposed \method is not only helpful for image classification, but also other visual tasks.

\subsection{Combining \method with SOTA methods}

Another line of research to address the long-tailed problem is the ensemble-based methods, such as MisLAS~\cite{zhong2021improving}, WD~\cite{alshammari2022long},  RIDE~\cite{wang2020long}, PaCo~\cite{cui2021parametric}, and BCL~\cite{zhu2022balanced}. 
Here we show that \method can also be leveraged to boost the performance of these methods. 

MisLAS 
use label-aware smoothing and shifted batch normalization to improve the second stage (the classifier learning) on the long-tailed recognition, while our method is to improve the first stage of representation learning. Thus, MisLAS could be combined with \method and TSC.
In Table~\ref{table:sota}, we report the results of combing MisLAS with both \method and TSC.
The results show that both \method and TSC improve the performance of  MisLAS and \method renders more performance boost than TSC.

Weight decay (WD)~\cite{alshammari2022long} applies a penalty on the magnitudes of the weights in a model, which is particularly prominent for larger weights, resulting in the acquisition of more balanced parameters.
WD and MaxNorm significantly enhances classification accuracy when applied to  a classifier~\cite{alshammari2022long}.
Table~\ref{table:sota} presents the accuracy of the classifier trained by WD and MaxNorm, based on different representations learned by \method and TSC.
The results demonstrate that SBCL improves the performance of the WD method~\cite{alshammari2022long} more effectively compared to TSC.

To implement \method with RIDE which incorporates multiple models in a multi-expert framework, we follow~\cite{li2021targeted} to simply replace the feature extractor on stage-1 training in RIDE with that trained with \method and keep the stage-2 routing training unchange.
As shown in Table \ref{table:sota}, applying \method to RIDE improves its performance with a significant
gap,  outperforms the combination of TSC and RIDE on all different number of experts.

PaCo and BCL propose new variants of supervised contrastive loss and jointly train both the proposed loss and classification loss to improve long-tail recognition, while we focus on the two stage pipeline, especially the first stage of representation learning.
In this experiment, we show that using models pretrained with both TSC and SBCL as initialization could improve the performance of both PaCo and BCL. The results can be found in Table~\ref{table:sota}, and we can see that SBCL renders larger performance gain than TSC. The improvement over PaCo and BCL  
sheds lights on potential future work to evaluate the combination of multiple techniques of long-tail recognition to achieve new SOTA results.

\begin{table}[h]
\centering
\caption{Performance of the combination of \method and SOTA ensemble-based methods
 with ResNet-50~\cite{he2016deep} on ImageNet-LT.
$\dag$ denotes the results of our re-implemented TSC.
}
\small
\label{table:sota}
\begin{tabular}{l|c|c|c|c}
\toprule[1pt]
Method  & Many  &  Medium &  Few  &  All    \\ 
\hline
MiSLAS~\cite{zhong2021improving}  &-   &- &-&52.7 \\
WD \& Max~\cite{alshammari2022long}&62.5 &50.4 &41.5 &53.9\\
RIDE~\cite{wang2020long} (3 experts) & 66.2   &51.7 &34.9&54.9  \\
PaCo~\cite{cui2021parametric} &64.4 &55.7 &33.7 &56.0 \\
BCL~\cite{zhu2022balanced}&67.9 &54.2 &36.6 &57.1\\

\hline
TSC+ MiSLAS &63.7 &50.5&36.0  &53.6\\
TSC+WD \& Max&63.8 &51.9 &41.6 &55.1\\
TSC+ RIDE (3 experts) &69.1 &51.7 &36.7 &56.3\\
TSC+ RIDE (3 experts)$\dag$ &68.6 &51.4 &36.0 &55.9\\
TSC+ PaCo &66.4 &55.8 &35.7 &57.1\\
TSC+ BCL &69.0 &56.3 &37.8 &58.7\\
\hline
\method+ MiSLAS&64.1 &52.0 &36.4&54.5\\
\method+WD \& Max&65.3  &52.6 &\textbf{42.7} &56.1\\
\method+ RIDE (3 experts) &69.2&52.4 &36.9 &56.8 \\
\method+ PaCo &66.9 &56.1 &38.4&57.9 \\
\method+ BCL &\textbf{70.0}&\textbf{56.9}&38.5&\textbf{59.5} \\
\bottomrule[1pt]
\end{tabular}
\end{table}


\begin{table}[h]
\centering
\caption{Ablation study on different components of \method.}
\begin{small}
\label{table:ablation}
\resizebox{1.\columnwidth}{!}{\begin{tabular}{ccc|c|c|c}
\toprule[1pt]
Warm-up & Adaptive cluster & Dynamic temperature  &\multicolumn{3}{c}{CIFAR-100-LT} \\
\hline
\multicolumn{3}{c|}{Imbalance Ratio}   & 100& 50 &10 \\ 
\hline
&$\checkmark$& $\checkmark$  &44.0 &47.9& 57.2 \\
$\checkmark$& &$\checkmark$  &43.8&47.2&56.5\\
$\checkmark$&$\checkmark$ &  &43.8 &47.8 &57.0\\
$\checkmark$&$\checkmark$ &$\checkmark$ & \textbf{44.9} & \textbf{48.7} & \textbf{57.9}\\
\bottomrule[1pt]
\end{tabular}}
\end{small}
\end{table}

\vspace{-1.0em}
\subsection{Ablation studies} \label{sec:abl}


\paragraph{Warm-up.} 
As mentioned in Section \ref{warm up}, we train the feature extractor for several epochs using ordinary SCL or KCL as warm-up stage.
As shown in Table \ref{table:ablation}, such a warm-up stage is beneficial since the performance drops when we remove the warm-up stage. 
This is likely because at the early stage of training, the extracted feature is not well-trained and the cluster assignment could be noisy and ineffective, hindering the efficacy of SBCL.


\paragraph{Adaptive clustering.}
We are also curious about the efficacy of adaptive clustering and thus present the performance of SBCL with clustering only once and fixing cluster assignments during training.
As shown in Table \ref{table:ablation}, without adaptive clustering, the performance decreases in all cases.
The reason could be fixed cluster assignment is prone to noise when the model is not well-trained, while adaptive clustering would dynamically adjust the cluster assignments based on the current learned model, which is supposed to become better as the training proceeds.

\begin{figure*}[t]
\centering
\begin{subfigure}{0.45\textwidth}
\centering
\includegraphics[scale=0.45]{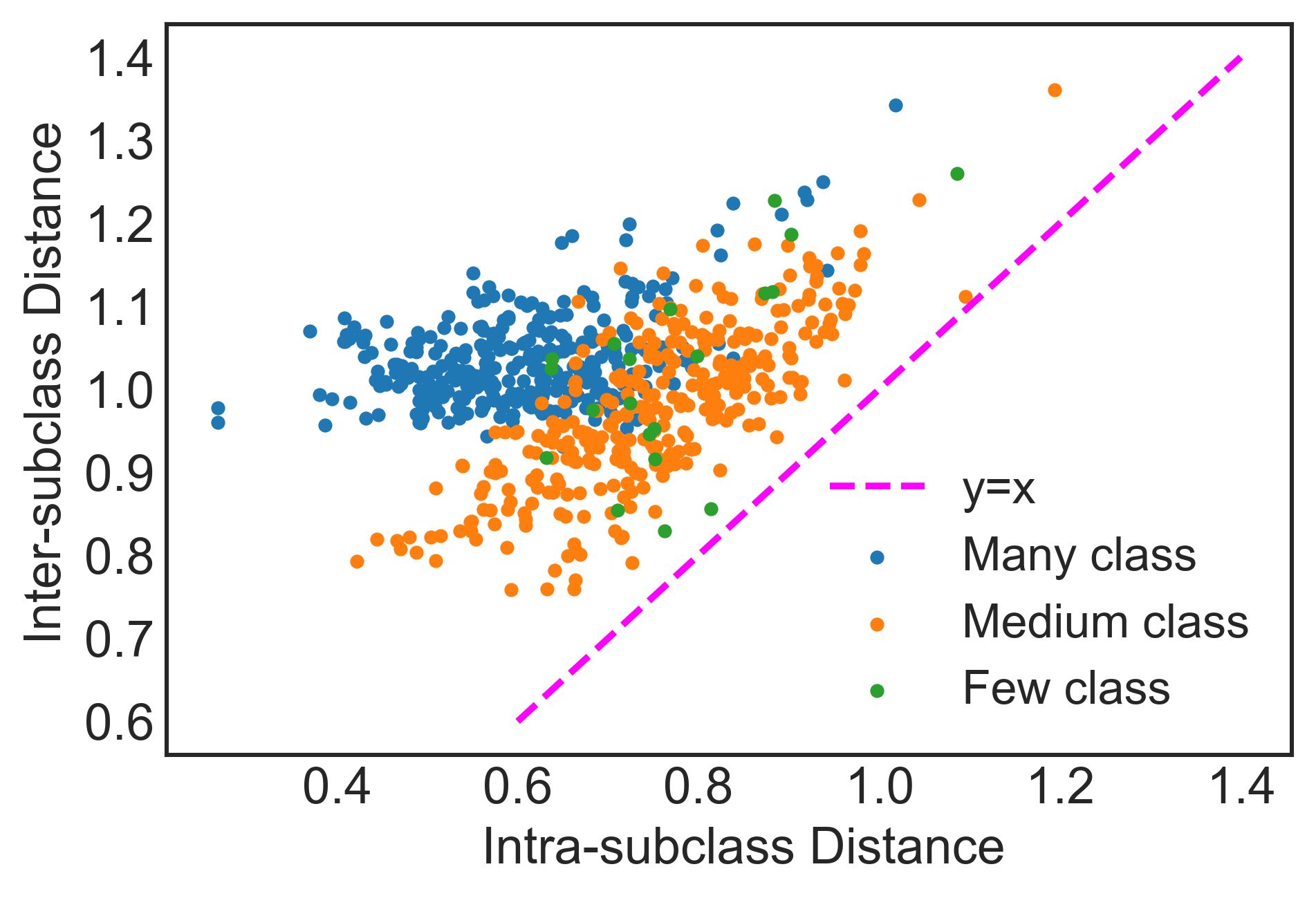}
\caption{Feature distance in subclasses.}
\label{fig2:1}
\end{subfigure}
\begin{subfigure}{0.45\textwidth}
\centering
\includegraphics[scale=0.45]{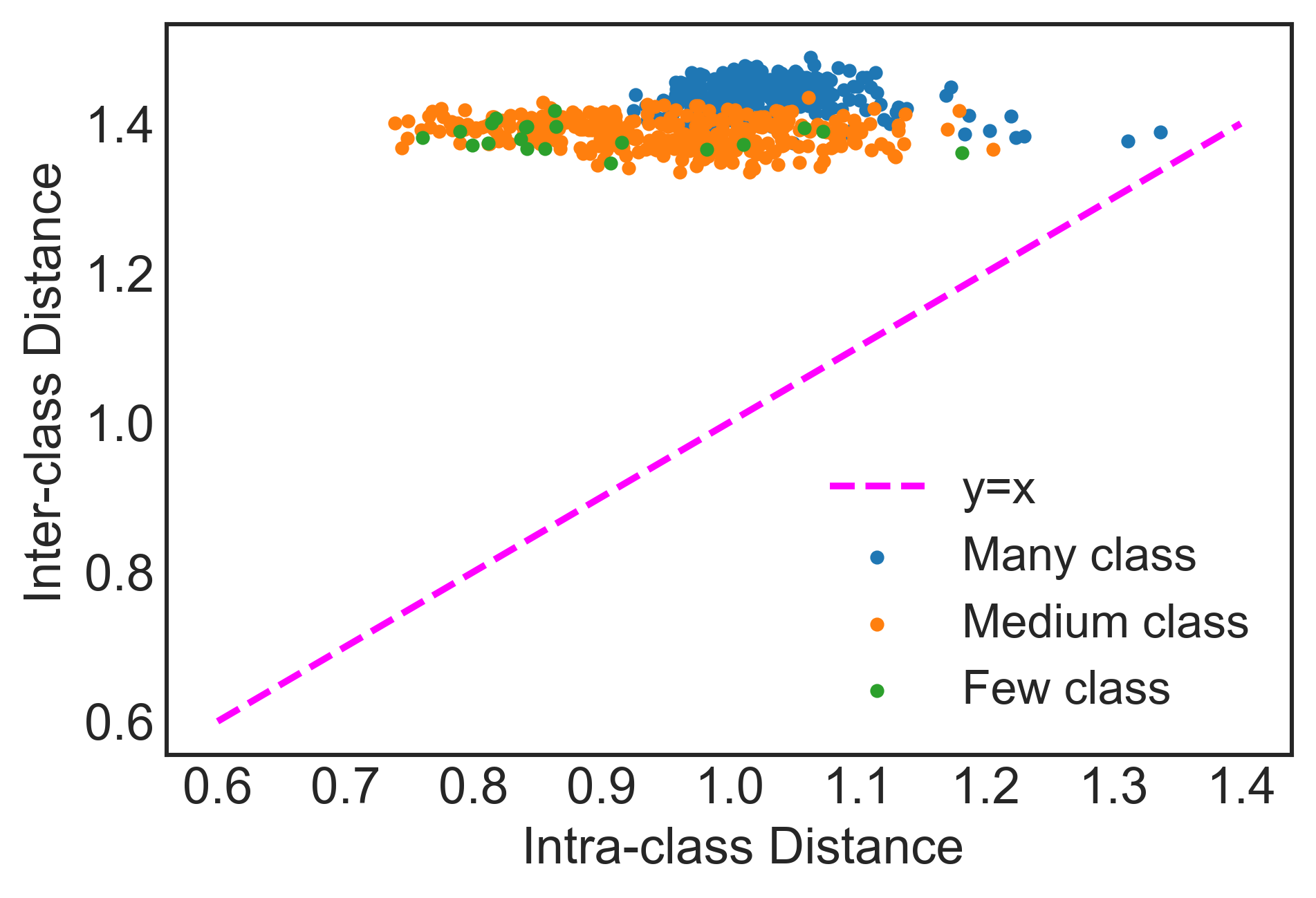}
\caption{Feature distance in classes.}
\label{fig2:2}
\end{subfigure}
\caption{Feature distance  of subclasses and classes on CIFAR-100-LT with imbalance ratio 100. We randomly sample instances from many-shot and medium-shot classes so that the size of each equals to that of few-shot classes.}
\label{fig_distance}
\end{figure*}

\paragraph{Dynamic temperature.}
In Table \ref{table:ablation}, we also study the effectiveness of dynamic temperature (Section~\ref{sec:bi}).
We remove the dynamic temperature and simply set $\tau_2>\tau_1$ following~\cite{rince2022AAAI}. 
With the fixed temperature $\tau_2$, the performance of \method is significantly worse than that with dynamic temperature.
We speculate that this is because 
dynamic temperature could help prevent the instances of a class to form overly large or small cluster in the feature space and therefore lead to better learned representations. Additionally, to evaluate the impact of dynamic temperature on other
baselines, we apply the dynamic temperature on TSC, as reported in Appendix~\ref{sec:dynamic temperature}.
In Appendix~\ref{app:time}, we also  provide an additional analysis of other hyperparametets on \method.

\subsection{Illustrate the superiority of \method}  \label{app:illustration}
In this section, we present an  example that explicitly demonstrates the superiority of SBCL compared to other baselines. 
We fix the features learned from  previous approaches  and \method on the CIFAR-100-LT with imbalance ratio 100, and use these features for classification task. Table~\ref{tab:acce} shows that SBCL achieves the best performance. 

\vspace{-1.0em}
\begin{table}[h]
\centering
\caption{Accuracy(\%) of SBCL and other baselines on CIFAR-100-LT with imbalanced ratio 100.}
\begin{small}
\label{tab:acce}
\begin{tabular}{c|c|c|c|c|c}
\hline
Method  &CE&SCL&KCL&TSC&SBCL \\
\hline
ACC(\%)&38.3&36.6&37.8&38.7&\textbf{39.9}\\
\hline
\end{tabular}
\end{small}
\end{table}

\subsection{Feature distribution of SBCL}

To analyze the representation learned by \method, we firstly define the euclidean distance between a given sample and other samples from the same/different classes as intra/inter-class distance. Concretely, the euclidean distance between a sample $z_i$ and a set $S$ is defined as $\mathbf{D}(z_i, S) = \frac{1}{|S|}\sum_{z_j \in S} \Vert z_i-z_j \Vert_2$. Then, the intra- and inter-class distance of sample $z_i$ can be defined as $\mathbf{D}(z_i, P_i)$ and $\mathbf{D}(z_i, \mathcal{D}/P_i)$ separately; and the intra- and inter-subclass distance of sample $z_i$ can be defined as $\mathbf{D}(z_i, M_i)$ and $\mathbf{D}(z_i, P_i/M_i)$ separately.
Figure~\ref{fig_distance} shows the average distance between features learned by SBCL in subclasses and classes on the CIFAR-100-LT.
As shown in Figure \ref{fig2:1}, 
the distance between samples from the
same subclass is less than those from the same class but different subclasses. 
Meanwhile, in Figure \ref{fig2:2}, 
 the inter-class distance consistently exceeds the intra-class distance, indicating clear separation between features from different classes. 
 Additionally, the inter-class distance remains stable, which suggests that  each class is uniformly distributed  on a hypersphere.
The results in all indicate that the two-layer class hierarchy is successfully captured and feature distribution achieves the core idea of \method. 
The specific values of feature distance are displayed in Appendix~\ref{app:distance}.


Figure~\ref{fig_class} 
visualizes the feature distribution of subclasses in a head class and head classes in CIFAR-100-LT with imbalance ratio 100 trained by \method using t-SNE~\cite{van2008visualizing}. 
The representations
learned by \method form some separated clusters, which be aligned with subclass labels and class labels.
This suggests that \method can learn an embedding space where features from same subclasses/classes are pulled closely and features from different subclasses/class are separated.

\begin{figure*}[h]
\centering
\begin{subfigure}{0.45\textwidth}
\centering
\includegraphics[scale=0.3]{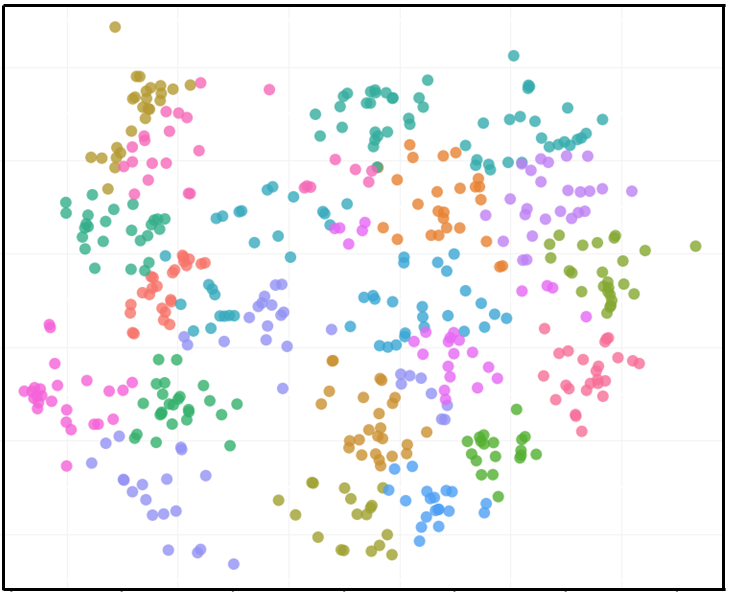}
\caption{Distribution of subclasses on  a head class.}
\label{fig2:subclass distribution}
\end{subfigure}
\begin{subfigure}{0.45\textwidth}
\centering
\includegraphics[scale=0.3]{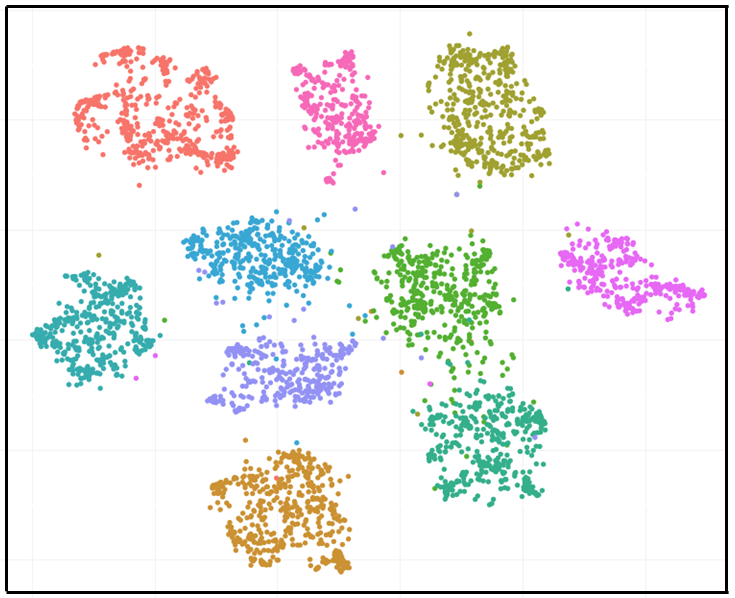}
\caption{Distribution of head classes on the  dataset.}
\label{fig2:class distribution}
\end{subfigure}
\caption{Feature distribution  of subclasses and classes on CIFAR-100-LT with imbalance ratio 100. Color represents subclasses/classes.}
\label{fig_class}
\end{figure*}

\subsection{Variation of per-class weight during training}

 Weight decay~\cite{alshammari2022long} achieves superior performance by effectively tuning weight decay during training. This study highlights the significance of regularizing network weights to become balanced for long-tailed recognition. Therefore, it is crucial to investigate whether \method can help the linear classifier to learn balanced weights through the training process.

\begin{figure}[h]
    \centering
    \includegraphics[scale=0.5]{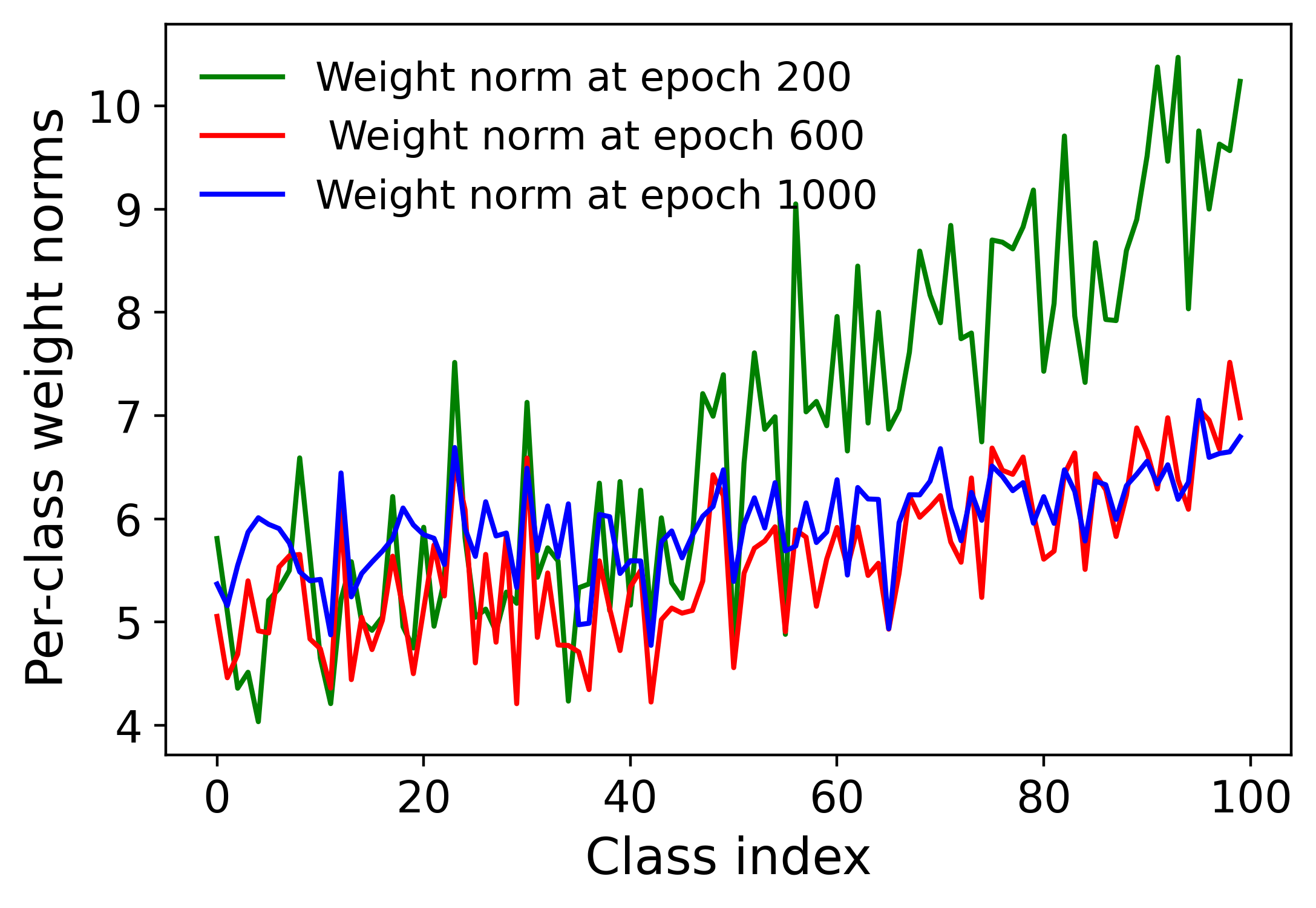}
    \caption{Change in weight norm of the linear classifier based on the representations trained by SBCL on CIFAR-100-LT with imbalance ratio 100 during training.}
    \label{fig:weight_norm}
    \vspace{-2.0em}
\end{figure}

Figure~\ref{fig:weight_norm} shows the per-class weight norm of a linear classifier trained on top of features learned by \method in different training stages on CIFAR-100-LT. 
From the figure, we can see that as the training proceeds, the per-class weight norm becomes  balanced even when training the linear classifier, the original cross-entropy loss is used.

\section{Related work}

Traditional methods for handling long-tailed recognition problem includes re-sampling and re-weighting.
There are roughly two types of re-sampling techniques: over-sampling the minority classes~\cite{shen2016relay,zhong2016towards,buda2018systematic,byrd2019effect} and under-sampling the frequent classes~\cite{he2009learning,japkowicz2002class,buda2018systematic}. 
The re-weighting techniques assign adaptive weights for different classes or even different samples. 
The vanilla scheme re-weights classes proportionally to the inverse of their frequency~\cite{huang2016learning,huang2019deep,wang2017learning}.
For class-level re-weighting methods, many loss functions including CB loss~\cite{cui2019class}, LDAM loss~\cite{cao2019learning} and Balanced softmax loss~\cite{ren2020balanced} were recently proposed, while instance-level re-weighting methods include Focal loss~\cite{lin2017focal} and Influence-balanced loss~\cite{park2021influence}.
Recently, two-stage algorithms have achieved remarkable performance for long-tailed recognition, such as classifier re-training (cRT)~\cite{kang2019decoupling}, learnable weight scaling (LWS)~\cite{kang2019decoupling}, and Mixup
Shifted Label-Aware Smoothing model (MiSLAS)~\cite{zhong2021improving}.
Meanwhile, bilateral branch network (BBN)~\cite{zhou2020bbn} uses an
additional network branch for re-balancing.
RIDE~\cite{wang2020long} use multiple branches named experts, each learning to specialize in the entire classes.
LADE~\cite{hong2021disentangling} assumes the prior of test class distributions is available and accordingly post-adjust model predictions. 
PaCo~\cite{cui2021parametric} applies parametric class-wise learnable centers to rebalance in contrastive learning.
BCL~\cite{zhu2022balanced} proposes a multi-branch framework to achieve class-averaging and class-complement in the training process.

To boost the performance of the two-stage algorithms, researchers have introduced supervised contrastive learning~\cite{khosla2020supervised} to the first feature-learning stage and proposed $k$-positive contrastive loss (KCL)~\cite{kang2020exploring} and targeted supervised contrastive learning (TSC)~\cite{li2021targeted}.
While achieving the state-of-the-art performance, these methods inject class-balance in the contrastive learning objective, inevitably leading to instance-imbalance during training.
In this work, we instead propose to achieve both subclass- and instance-balance in the contrastive learning object.
Our method is also related to recent studies of clustering-based deep unsupervised learning ~\cite{dosovitskiy2014discriminative,xie2016unsupervised,liao2016learning,yang2016joint,caron2018deep,caron2020unsupervised}, especially those that leverage contrastive learning~\cite{li2020contrastive,PCL,wang2021unsupervised,guo2022hcsc}.
However, they target at general unsupervised representation learning scenario, while our method is tailored for long-tailed recognition where the training data is immensely class-imbalanced.

\section{Conclusion}
\label{sec:conclusion}

In this paper, we introduced Subclass-balancing Contrastive Learning (SBCL) for long-tailed recognition. 
It breaks down the head classes into multiple semantically-coherent subclasses via subclass-balancing adaptive clustering and incorporates a bi-granularity contrastive loss that encourages both subclass- and instance-balance. 
Extensive experiments on multiple datasets demonstrate that \method achieves state-of-the-art single-model performance on benchmark datasets for long-tailed recognition.

\clearpage

\nocite{*}

\bibliographystyle{ieee_fullname}
\bibliography{r.bib}

\clearpage

\appendix
\newpage
\section{APPENDIX}

\subsection{Additional experiment results}

\begin{figure*}[t]
    \centering
    \includegraphics[scale=0.25]{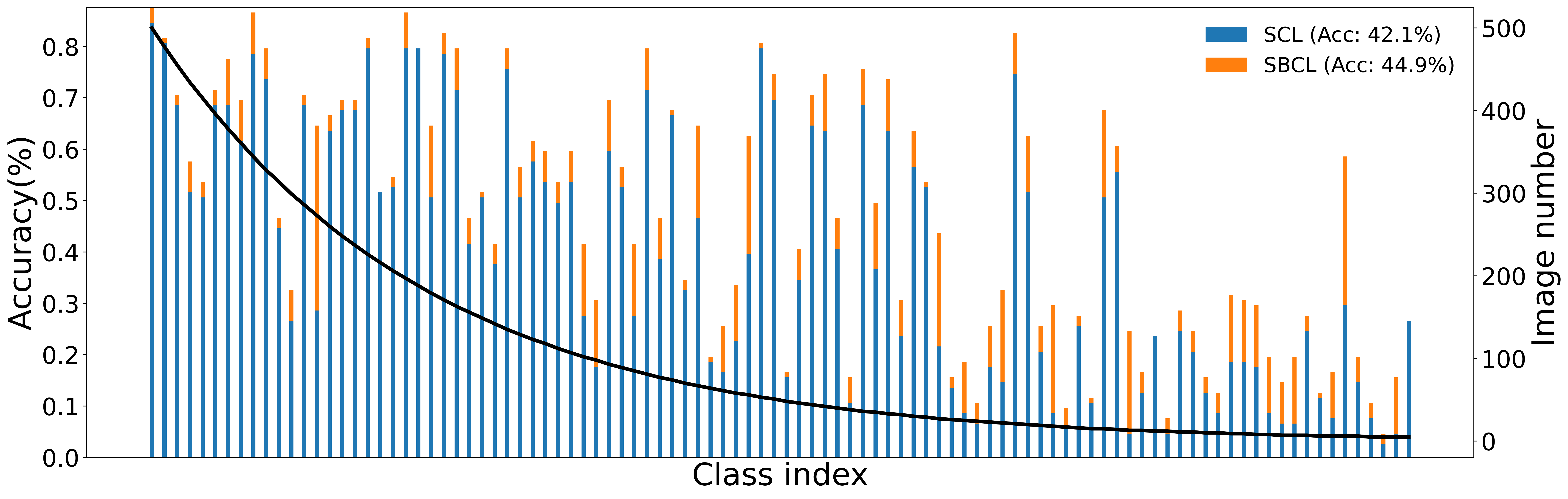}
    \caption{Accuracy of each classes on CIFAR100-LT. The black line is the class distribution, and the classes in the left part are head classes while those in the right part are tail classes.}
    \label{fig:acc_cifar}
\end{figure*}

\paragraph{Accuracy of each class on CIFAR-100-LT.}

We visualize the accuracy of each class of both SCL and SBCL on CIFAR-100-LT with imbalance ratio 100  (Figure~\ref{fig:acc_cifar}). 
From the results, we can see that \method improves performance on tail classes over SCL without the expense of the perforamnce of the head classes.




\paragraph{Selection of negative instances in SBCL.}

Our proposed loss in Eq.~\ref{eq:loss} consists of two supervised contrastive losses with subclass and class labels respectively.
The first term regards instances in different subclasses as negative instead of these in different classes; In Table~\ref{table: select negative}, we show that such design choice leads to better performance than using instances in different classes as negative, which  illustrates the effectiveness of exploiting the rich semantic in head classses.

\begin{table}[h]
\centering
\caption{Different selection of negative instances in SBCL on CIFAR-100-LT with different imbalance ratio.
The ‘Class label‘ row  are the first term of loss function is constructed by class label and the 'Subclass label' row are subclass label.}
\begin{small}
\label{table: select negative}
\begin{tabular}{c|c|c|c}
\toprule[1pt]
\multirow{2}{*}{Negative samples} &\multicolumn{3}{c}{Imbalance Ratio} \\
\cline{2-4}
  & 100& 50 &10 \\
\hline
 Class label&43.7&47.6 &56.8\\
 Subclass label&\textbf{44.9}&\textbf{48.7}&\textbf{57.9}\\
\bottomrule[1pt]
\end{tabular}
\end{small}
\end{table}

\paragraph{Analysis of feature distribution.}
\label{app:distance}

To analyze the representation learned by \method, we firstly define the euclidean distance between a given sample and other samples from the same/different classes as intra/inter-class distance.
Concretely, the euclidean distance between a sample $z_i$ and a set $S$ is defined as $\mathbf{D}(z_i, S) = \frac{1}{|S|}\sum_{z_j \in S} \Vert z_i-z_j \Vert_2$. Then, the intra- and inter-class distance of sample $z_i$ can be defined as $\mathbf{D}(z_i, P_i)$ and $\mathbf{D}(z_i, \mathcal{D}/P_i)$ separately; and the intra- and inter-subclass distance of sample $z_i$ can be defined as $\mathbf{D}(z_i, M_i)$ and $\mathbf{D}(z_i, P_i/M_i)$ separately.

To leverage instance semantic coherence to balance the feature space, we expect 
instances of high semantic coherence to form a more concentrated cluster than other instances in the same class.
So, we embed the subclass-balancing adaptive clustering strategy on SBCL to illustrate this on CIFAR-100-LT with imbalance ratio 100.
In Table \ref{table:class}, we report the intra-subclass/inter-subclass distance on different splits. 
The results show that \method achieves to concentrate instances from the same subclass and pulls instances from different subclasses away on all splits.

\method aims at learning a compact representation space, in which representations from different classes are far from each other and the feature space spanned by representations of each class is invariant to the long-tailed distribution.
The average intra/inter-class distance are summarized in Table \ref{table:class} and the distances of different groups are reported separately.
The results also show  \method clears the
boundary of feature distribution on the different class splits.

\begin{table}[h]
\centering
\caption{Average intra/inter-subclass and intra/inter-class distance of features learned by SBCL.}
\begin{small}
\label{table:class}
\begin{tabular}{l|c|c|c|c}
\toprule[1pt]
Distance   & Many  &  Medium &  Few  &  All    \\
\hline
Intra-subclass  &0.68&0.76 &0.87 &0.70\\
Inter-subclass &1.02&0.99 &0.89  &1.01\\
\bottomrule[1pt]
Intra-class  &1.00 &0.94 &0.88 &0.99\\
Inter-class  &1.39&1.38  &1.37&1.39\\
\bottomrule[1pt]
\end{tabular}
\end{small}
\end{table}


\paragraph{Subclass-balancing adaptive clustering.}

We propose an noval $K$-means algorithm aimed at achieving balanced intra-cluster sample quantities.

In the initial step, we adopt a strategy to select cluster center points that are maximally distant from each other. This ensures an optimal distribution of initial cluster centers~\cite{arthur2007k}.

Next, in the assignment step, we calculate the similarity between each sample point and the selected cluster centers. The sample points are then assigned to the cluster centers in descending order of their similarity scores. However, we introduce a modification to this step by incorporating a constraint on the maximum number of samples assigned to a given cluster center. Once this threshold is reached, the cluster center is not eligible to receive any samples.

In the update step, we revise the cluster centers which are determined as the average values of the samples allocated to this cluster.

We continue to iterate through the assignment and update steps until we satisfy the termination condition, which is achieving a predetermined number of iterations ($M$). This iterative process facilitates a balanced distribution of samples within each cluster, leading to release the long-tailed phenomena.

\paragraph{Hyperparameter analysis on CIFAR-100-LT.} \label{app: sample number}


\begin{figure*}[t]
	\centering
	\begin{subfigure}{0.46\linewidth}
		\centering
		\includegraphics[scale = 0.46]{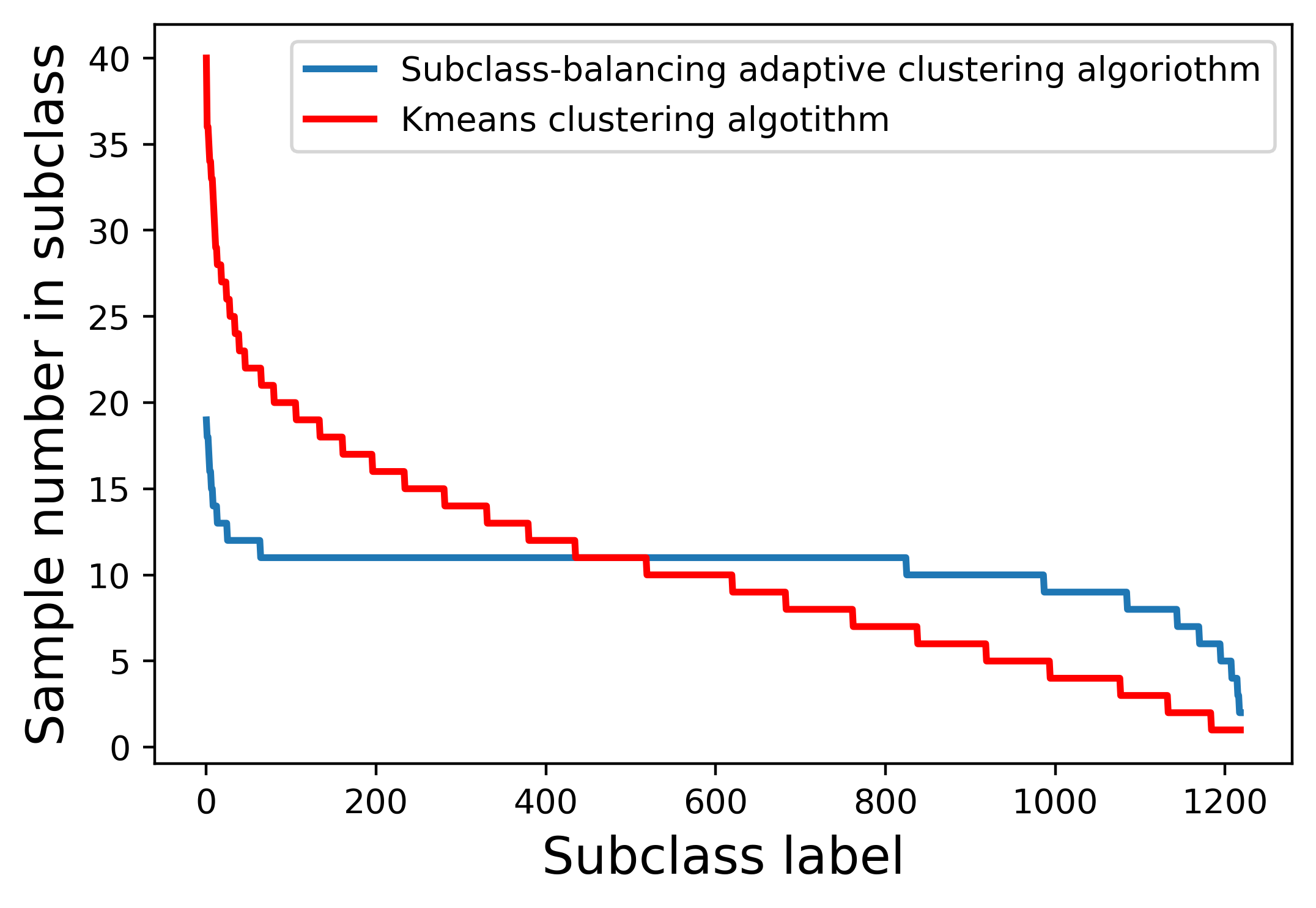}
		\caption{}
		\label{fig:cifarlongtail}
	\end{subfigure}
	\centering
	\begin{subfigure}{0.46\linewidth}
		\centering
		\includegraphics[scale = 0.46]{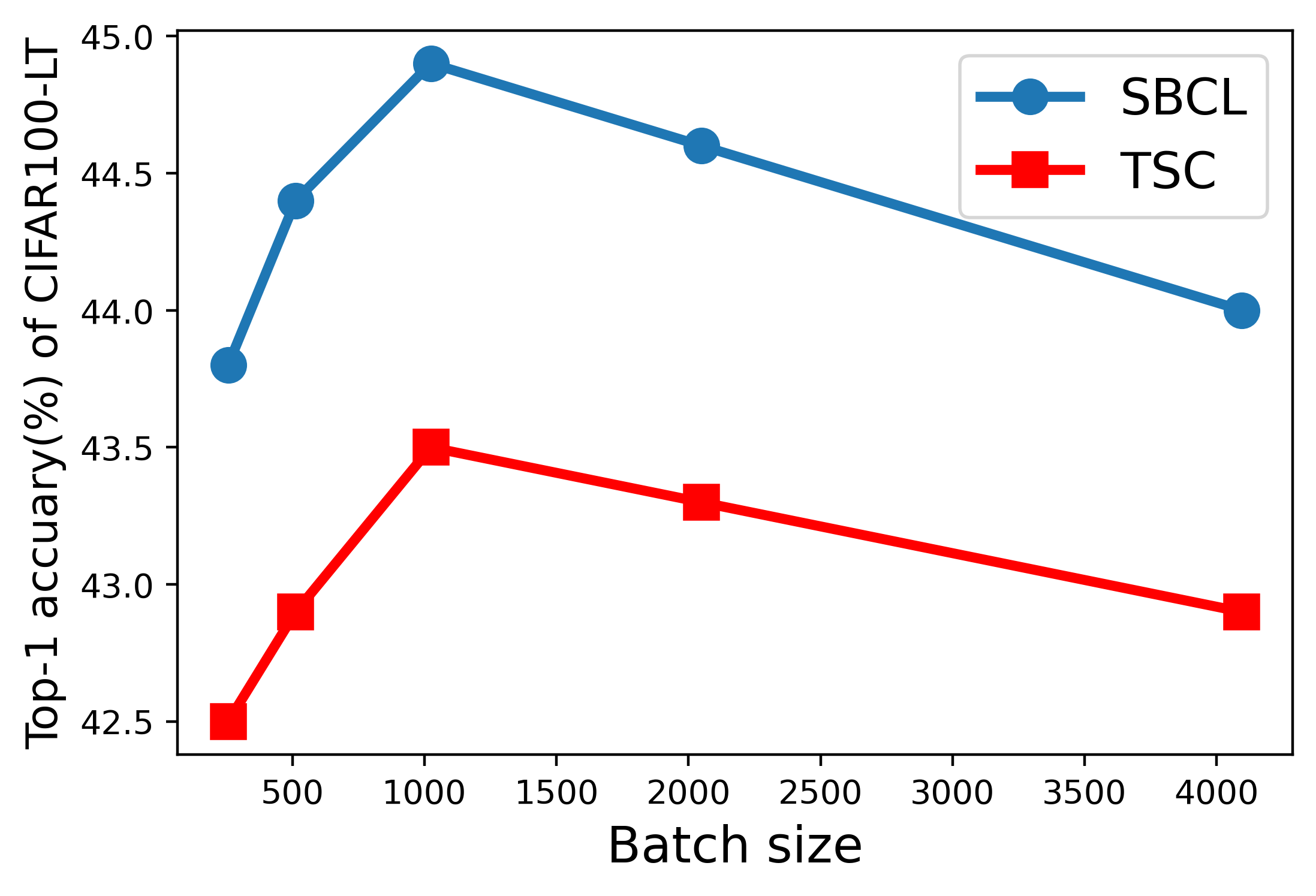}
		\caption{}
		\label{fig: batch size}
	\end{subfigure}
	\centering
	\begin{subfigure}{0.46\linewidth}
		\centering
		\includegraphics[scale = 0.46]{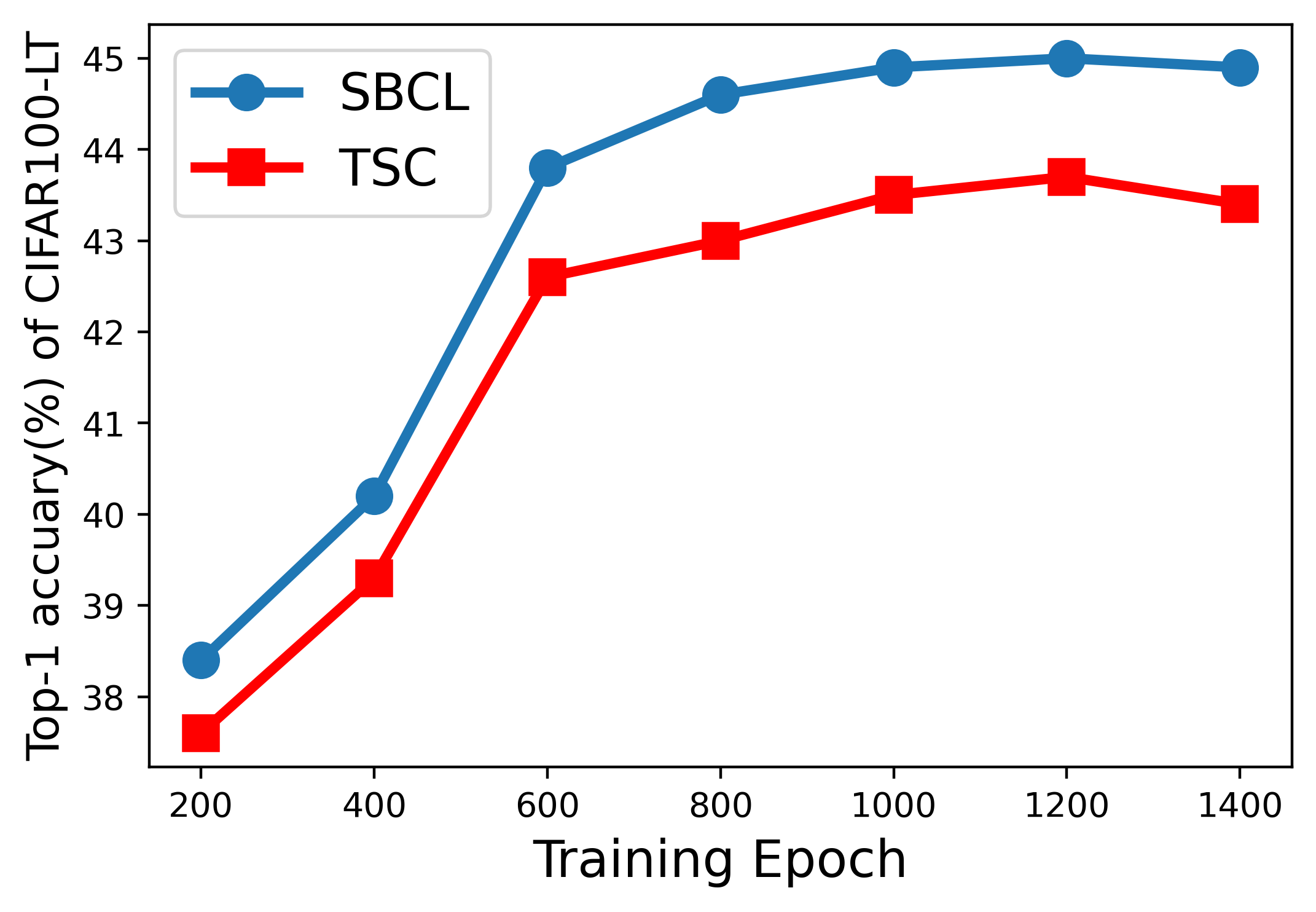}
		\caption{}
		\label{fig: epoch}
	\end{subfigure}
	\begin{subfigure}{0.46\linewidth}
		\centering
		\includegraphics[scale = 0.46]{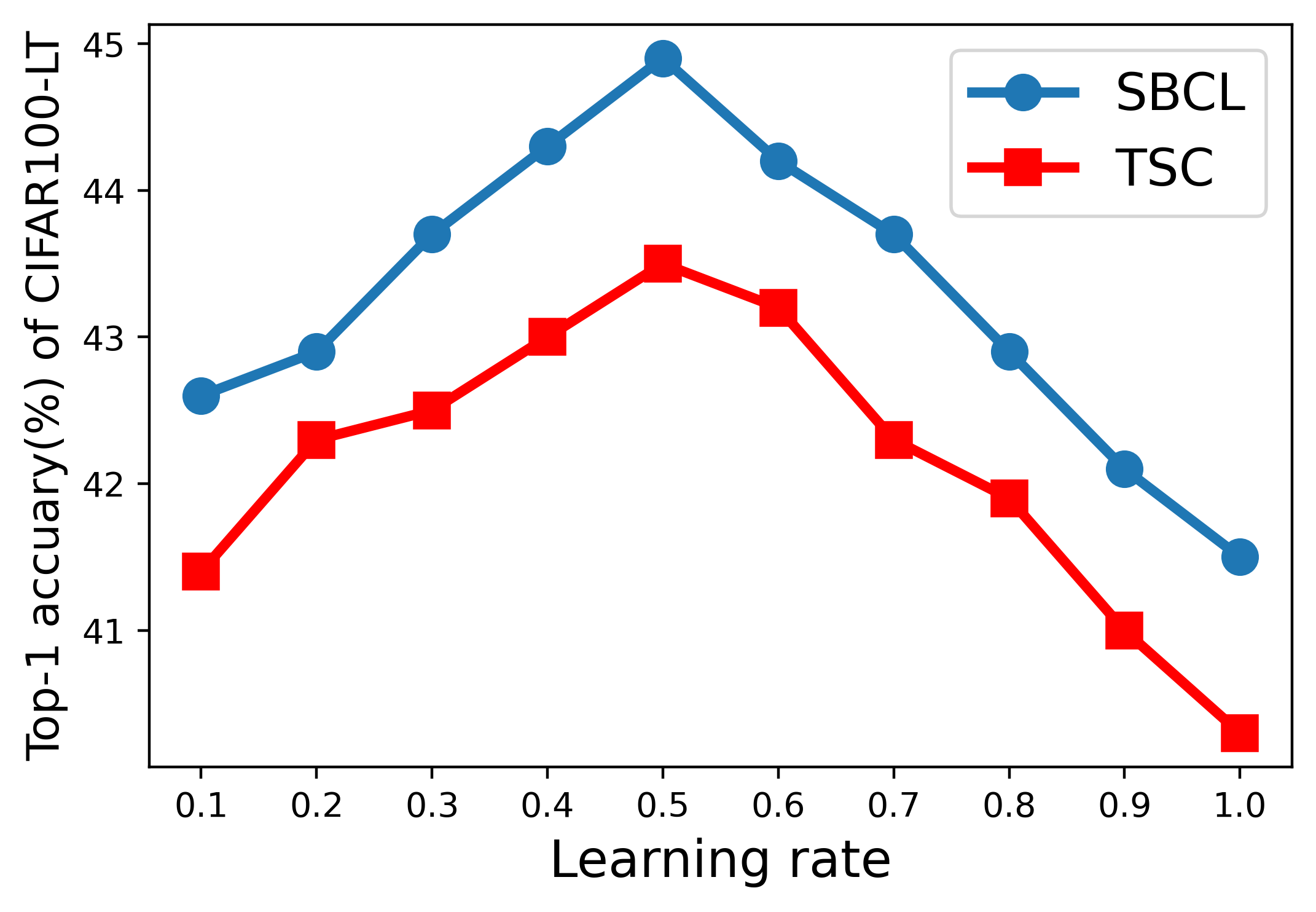}
		\caption{}
		\label{fig: learning rate}
	\end{subfigure}
	\caption{\textbf{Analysis of SBCL as a loss function of different hyperparameters on CIFAR-100-LT with imbalance ratio 100.}  (a): Sample number in clusters with different cluster algorithm. (b): Top-1 accuracy of SBCL/TSC as a function of different batch size. (c): Top-1 accuracy of SBCL/TSC as a function of different pretraining epochs. (d): Top-1 accuracy of SBCL/TSC as a function of different learning rates.}
	\label{da_chutian}
\end{figure*}

\begin{table*}[h]
\centering
\caption{Distribution of sample number in subclass on CIFAR-100-LT with imbalance ratio 100.}
\small
\label{table:imbalance}

\begin{tabular}{lccccc}
\toprule[1pt]
Dataset & Max  & Min  &  Average & Std  &  Imbalance ratio(Max/Min)    \\ 
\hline
Kmean clustering algorithm &40 & 1  &10.34 &6.27 &40 \\
Subclass-balancing adaptive clustering algorithm &19 & 2  &10.34 &1.60 &9.5 \\
\bottomrule[1pt]
\end{tabular}
\end{table*}

Figure~\ref{fig:cifarlongtail} and Table \ref{table:imbalance} show the distribute situation of  sample number in subclasses obtained by different cluster algorithms on CIFAR-100-LT with imbalance ratio 100.
For Kmean cluster algorithm, the imbalance phenomenon of subclasses is obvious.
When using our proposed cluster algorithm, the imbalance ratio of sample number in subclasses deceases from 40 to 9.5.
And the standard deviation of sample number on CIFAR-100-LT is relatively small, which denotes the number of samples in most subclasses keeps stable in a certain range.

Figure~\ref{fig: batch size}  shows the impact of batch size on SBCL/TSC.
We find that larger batch sizes have a significant advantage over the smaller ones. 
This is because larger batch sizes provide more negative examples to facilitate convergence. However, the over-large batch size hurts the model performance. And SBCL and TSC are equally sensitive to batch size on CIFAR-100-LT.

Figure~\ref{fig: epoch}  shows the curve of the accuracy of SBCL/TSC  vs. the number of training epochs. 
From the curve, we can see that the performance of \method and TSC both converge after 800 epochs. When the model is trained with SBCL over 600 epochs, its performance already exceeds TSC.

In Figure~\ref{fig: learning rate}, we display the performance of SBCL with different learning rates on
CIFAR-100-LT with imbalance ratio 100.
As shown in the figure, the learning rate has significant impact on the performance, and we set the learning rate as 0.5 for CIFAR-100-LT.


\paragraph{Combining TSC with dynamic temperature.}
\label{sec:dynamic temperature}

According to Table \ref{table:ablation}, dynamic temperature effectively contributes to the improvement of accuracy.
We also add the dynamic temperature to the second term of TSC~\cite{li2021targeted} and the experiment results are shown in Table ~\ref{table: technique}.
However, the improvement of the dynamic temperature on TSC is less significant than that on our method, which is reasonable because we introduce dynamic temperature for the loss to distinguish between class and subclass, while TSC does not have subclass and therefore the dynamic temperature is less effective. 

\begin{table*}[h]
\centering
\caption{Combination of TSC and SBCL with dynamic temperature.}
\small
\label{table: technique}

\begin{tabular}{c|c|c|c|c|c|c}
\toprule[1pt]
\multirow{2}{*}{Dynamic temperature}  &\multicolumn{6}{c}{CIFAR-100-LT} \\
\cline{2-7}
  &\multicolumn{3}{c|}{TSC} &\multicolumn{3}{c}{SBCL}\\
\hline
Imbalance Ratio  & 100& 50 &10 & 100& 50 &10 \\
\hline
&43.5&47.6 &58.7&43.8 &47.8 &57.0 \\
 $\checkmark$  &43.9\textcolor{red}{(+0.4)} &48.0\textcolor{red}{(+0.4)} &59.2\textcolor{red}{(+0.5)} & 44.9\textcolor{red}{(+1.1)} &48.7\textcolor{red}{(+0.9)}& 57.9\textcolor{red}{(+0.9)}\\
\bottomrule[1pt]
\end{tabular}
\end{table*}

\paragraph{Time efficiency comparison of SBCL and TSC.}
\label{app:time}

\begin{table}[h]
\centering
\caption{Computing cost (GPU hours) on CIFAR-100-LT dataset with imbalance ratio 100.}
\begin{small}
\label{tab:gpuhour}
\begin{tabular}{c|c|c}
\hline
Method  &TSC &SBCL \\
\hline
GPU hours &2 &2.4 \\
\hline
\end{tabular}
\end{small}
\end{table}

In Table~\ref{tab:gpuhour}, SBCL consumes more time, because we use a clustering algorithm on head classes to train SBCL on CIFAR-100-LT with imbalance ratio 100. However, we update the clustering results every few training iterations, and ultimately achieve better results than TSC. Therefore, we believe that the additional  small computational cost is worth the effort.

\paragraph{Warm-up on ImageNet-LT.}
Instead of using the SCL at the warm-up stage for CIFAR datasets, KCL is adopted for ImageNet-LT and iNaturalist 2018 datasets to warm up the feature extractor.
As Table \ref{table:warmimage} shows, warm-up phase makes feature extractor improve accuracy on all splits of ImageNet-LT.
This is because it prevents cluster assignment from feature random distribution at the beginning and avoids using the SCL to make the feature space dominated by the head class at the warm-up stage.

\begin{table}[http]
\centering
\caption{SBCL with and without warm-up stage on ImageNet-LT. 
}
\small
\label{table:warmimage}
\begin{tabular}{l|c|c|c|c}
\toprule[1pt]
Method & Many &Medium & Few &All \\ 
\hline
SBCL without warm-up&62.9 &49.6 &29.3 &52.0 \\

SBCL with warm-up &\textbf{63.8} &\textbf{51.3} &\textbf{31.2} &\textbf{53.4}  \\
\bottomrule[1pt]
\end{tabular}
\end{table}

\paragraph{Advantages of cluster validity.}
Actually, previous studies~\cite{kang2020exploring,li2021targeted} have proven that randomly sampling balanced instances as positive pairs  (such as KCL, TSC) is better than sampling all instances of the same class as positive pairs (such as SCL).
However, this strategy may destruct instance semantic coherence. 
In Table \ref{table:cluster strategy}, we replace the first team (regard subclasses as positive pairs) with the balanced positive sampling strategy (KCL) to prove this on ImageNet-LT. 
As the results show, subclass-balancing adaptive clustering strategy brings more improvement to \method than balanced positive sampling strategy. 
\begin{table}[http]
\centering
\small
\caption{Subclass-balancing adaptive clustering strategy  improves more than  balanced positive sampling strategy  on ImageNet-LT.}
\label{table:cluster strategy}
\begin{tabular}{l|c|c|c|c}
\toprule[1pt]
Method  & Many  &  Medium &  Few  &  All    \\ 
\hline
FCL &61.4   & 47.0 &28.2 &49.8  \\
KCL  & 62.4 &49.0& 29.5 &51.5 \\
TSC  & 63.5   &49.7&30.4  &52.4 \\
\hline
\method(KCL) &63.3 &49.5 &30.6&52.2 \\
\method  &\textbf{63.8} &\textbf{51.3} &\textbf{31.2} &\textbf{53.4}  \\
\bottomrule[1pt]
\end{tabular}
\label{table_MAP}
\end{table}

\paragraph{COCO object detection and instance segmentation.} \label{appendix:coco}

In this section, following the experiment setting in~\cite{he2020momentum}, we use Mask R-CNN~\cite{he2017mask} to conduct the object detection and instance segmentation experiments on COCO dataset. 
The schedule is the default 2× in~\cite{he2020momentum}.
Table~\ref{table:coco} shows the pretrained model trained by \method outperforms it learned with other contrastive learning for the downstream tasks.

\begin{table*}[h]
\centering
\caption{\textbf{Object detection and instance segmentation results on COCO dataset.} The representation model is trained on ImageNet and ImageNet-LT. We report results in bounding-box AP (AP$^{bb}$) and mask AP (AP$^{mk}$). }
\label{table:coco}
\small
\begin{tabular}{l|l|c|c|c|c|c|c}
\toprule[1pt]
\multicolumn{2}{c|}{\multirow{2}*{Method}} &\multicolumn{3}{c|}{ImageNet}  &\multicolumn{3}{c}{ImageNet-LT} \\
\cline{3-8}
\multicolumn{2}{c|}{~}& AP &AP$_{50}$   &AP$_{75}$ & AP&AP$_{50}$ &AP$_{75}$ \\ 
\hline
\multirow{6}*{AP$^{bb}$}
&random init. &35.6 &54.6 &38.2&35.6 &54.6 &38.2\\
&CE    & 40.1 &59.8 &43.3 &38.1 &57.4 &41.2  \\
  ~&CL~\cite{he2020momentum} &40.4 & 60.1 & 44.1 & 39.7 & 59.4 & 42.7 \\
  ~&KCL~\cite{kang2020exploring}& 40.8 & 60.6 & 44.0 & 39.4 & 59.1 & 42.6 \\
 \cline{2-8}
 ~&\method &\textbf{41.1}   &\textbf{60.8} &\textbf{44.2}&\textbf{40.0} &\textbf{59.6}&\textbf{43.0} \\\hline
\multirow{6}*{AP$^{mk}$}
&random init. &31.4 &51.5 &33.5&31.4 &51.5 &33.5\\
&CE    & 34.9 &56.6 &37.0 &33.3 &54.2 &35.4  \\
  ~&CL~\cite{he2020momentum} &35.1 & 56.9 & 37.6 & 34.7 & 56.1 & 37.1 \\
  ~&KCL~\cite{kang2020exploring}& 35.5 & 57.4 & 37.8 & 34.4 & 55.8 & 36.4 \\
 \cline{2-8}
 ~&\method &\textbf{35.7}   &\textbf{57.5} &\textbf{37.9}&\textbf{35.0} &\textbf{56.3} &\textbf{37.3} \\ 
\bottomrule[1pt]
\end{tabular}
\end{table*}

\paragraph{Hyperparameter studies.}
Here, we study the effect of hyperparameters $\beta$ and $\delta$. Note that $\beta$ controls the balance of two loss terms in Eq.~\ref{eq:loss} and $\delta$ determines  the lower bound of the cluster size in Eq.~\ref{eq:cluster}.
Specifically, on CIFAR-100-LT with imbalance ratio 100, we vary the values of $\beta$ from \{0.1, 0.2, 0.5, 0.8, 1.0, 2.0\} with $\delta=10$ and the value of $\delta$ from \{5, 10, 20, 30, 50, 100\} with $\beta=0.2$. The results are summarized in Table~\ref{tabel:beta}.
We observe that the smaller $\beta$ values (between 0.1 and 0.5) can achieve relatively good performance, with the best being 0.2.
This observation aligns with our intuition of emphasizing the subclass-level contrastive loss, because smaller $\beta$ is equivalent to putting more weights on the first term of Eq.~\ref{eq:loss}, which corresponds to the subclass-level contrastive.
For $\delta$, the values between 5 and 30 yield high accuracy, with the best being 10.
We can see that large $\delta$ values ($\delta=50,100$) lead to significant drop in performance.
We argue that this is because large $\delta$ value would result in subclasses that contain more instance than tail classes and therefore affect the subclass-balance, leading to suboptimal performance.  
In addition, smaller $\delta$ value ($\delta=5$) also causes performance drop; the reason could be small cluster size may let similar instance being assigned to different clusters and therefore affect the learned representations.
Therefore, we fix  $\beta =0.2$ and $\delta=10$ for all experiments.
\begin{table}[h]
\centering
\caption{Hyperparameter study of $\beta$ and $\delta$ on CIFAR-100-LT with imbalance ratio 100.}
\label{tabel:beta}
\small
\begin{tabular}{c|c|c|c|c|c|c}
\toprule[1pt]
 $\beta$  &  0.1     &  0.2  &  0.5 &  0.8   & 1.0  & 2.0  \\
 \hline
 ACC(\%) &44.6 &\textbf{44.9} &44.5 &44.1& 43.9& 42.1 \\
 \hline
 \hline
 $\delta$  & 5    &  10  &  20 &  30  & 50  & 100  \\
 \hline
 ACC(\%) &44.3 &\textbf{44.9}&44.6 &44.3 &42.9 &42.3 \\
\bottomrule[1pt]
\end{tabular}
\vspace{-10pt}
\end{table}

\begin{figure}[h]
    \centering
    \includegraphics[scale=0.22]{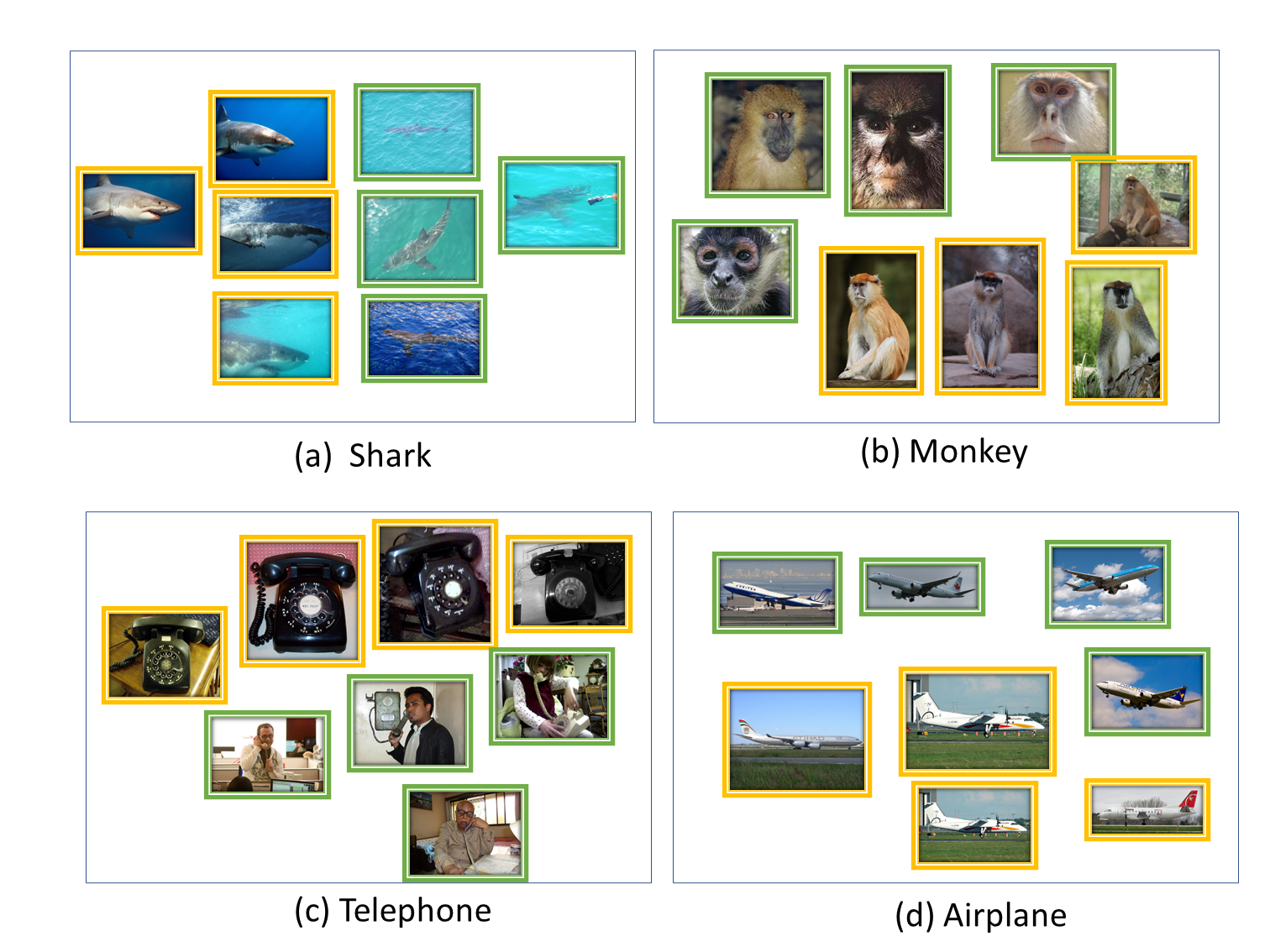}
    \caption{Visualization of subclasses generated by SBCL. Images with green and orange boarder are randomly drawn from different subclasses within the same classes. We can see that SBCL could produce semantically coherent subclasses.}
    \label{fig:clusteringpic}
\end{figure}

\paragraph{Visualization of generated clusters.}
In Figure~\ref{fig:clusteringpic}, we show the clustering results of ImageNet-LT training images generated by subclass-balancing adaptive clustering algorithm. 
From the results, we can see that the algorithm is able to find the subclasses with similar patterns, helping the model learn semantic coherent representations. For example, the two subclasses in the bottom-left are telephone with/without human.

\subsection{Additional information}

\paragraph{Benchmark datasets statistical information and Implementation Details.}
\label{apd:implementation_detail}
\begin{table}[h]
\small
\centering
\caption{Statistics of datasets. The imbalance ratio $\rho = n_1 / n_C$. }
\label{table:sum}
\resizebox{1.0\linewidth}{!}{
\begin{tabular}{lcccc}
\toprule[1pt]
Dataset &  classes &  training data & test data & imbalance ratio\\
\hline
CIFAR-100-LT & 100 & 50,000 &10,000 & $\left\{100,50,10\right\}$\\
ImageNet-LT & 1,000 & 115,846 & 50,000 & 256\\
iNaturalist 2018 & 8,142 &437,513 &24,426 &500\\
\bottomrule[1pt]
\end{tabular}
}
\end{table}

\begin{table*}[h]
\small
\centering
\caption{Hyperparameters used by different loss functions for benchmark datasets. The detailed hyperparameters of iNaturalist 2018 are the same as the ImageNet-LT.}
\label{table:hyper}
\setlength{\tabcolsep}{3.8mm}{
\begin{tabular}{r|cc|cc}
\toprule[1pt]
\multirow{2}*{Hyperparameters} & \multicolumn{2}{c|}{ImageNet-LT} & \multicolumn{2}{c}{CIFAR100-LT}\\
\cline{2-5}
~  & TSC & SBCL & TSC & SBCL\\
\hline
module & MoCo & MoCo & SimCLR & SimCLR \\
warm-up epoch & 200 & 200 & 0 & 10 \\
epoch  & 400 & 400 & 1000 & 1000 \\
batch size  & 256 & 256 & 1024 & 1024\\
learning rate  & 0.1 & 0.1 & 0.5 & 0.5\\
learning rate schedule  & cosine &cosine & cosine & cosine\\
memory size  & 65536 & 65536 & - & - \\
encoder momentum   & 0.999 &0.999 & - & -\\
feature dimension  &128 &128& 128 & 128\\
softmax temperature  &0.07 &0.07 & 0.1 & 0.1\\
$k$-positive number &6 &- & 4 & -\\
hyperparameter of  $\beta$ &0.2&0.2&0.2 & 0.2\\
hyperparameter of  $\delta$ 
 &- &20 & - & 10\\
 hyperparameter of  $\alpha$&- &10 & - & 10\\
\bottomrule[1pt]
\end{tabular}
}
\end{table*}

We summarize the statistical information of the three benchmark datasets in Table \ref{table:sum}.
Following~\cite{kang2019decoupling,kang2020exploring,li2021targeted}, we apply \method on the long-tailed recognition by using a two-stage training strategy: (i) train the representation with \method; (ii) learn a linear classifier on top of the fixed representation.
The training process is the same as TSC~\cite{li2021targeted}.
Thus,
we use TSC default hyperparameters and implementation details for the representation learning.
For CIFAR-100-LT dataset, all experiments are performed on 2 NVIDIA RTX 3090 GPUs.
For ImageNet-LT and iNaturalist 2018 datasets, we  perform the experiments on 8 NVIDIA RTX 3090 GPUs.
The detailed hyperparameters of TSC and SBCL are given in Table \ref{table:hyper}.

For the classify learning,  training the linear classifier strategy is the same with
TSC~\cite{li2021targeted}; so, we use TSC default hyperparameters and implementation details for the classifier learning. For the detect model learning, we follow MoCo~\cite{he2020momentum} to adopot the same setting, hyperparameters and evolution metrics with  R50-C4 backbone.
For Pascal VOC dataset, we train Faster R-CNN~\cite{ren2015faster} on  VOC07+12  and evaluate  on the test set of VOC07. 
For COCO dataset, we train Mask R-CNN~\cite{he2017mask} on train2017 set and evaluate on val2017 set.

\paragraph{Limitations.}
\method has some limitations. 
First, clustering the head class in \method takes a long time on the training phase, especially for ImageNet-LT and iNaturalist 2018.
Second, \method requires knowing the number of samples in each class to decide the cluster number; so, it is not applicable to problems where the number of samples is unknown. 

\paragraph{Social impacts.}
This work aims to propose a novel representation learning to help people resolve the bias in the real world data recognition, which might has positive social impact.
We do not foresee any form of negative social impact induced by our work.

\paragraph{Privacy information in data.}
All datesets we used in the experiment are public.
The datasets only include the pictures, which most are  animals and plants. No private information is included.

\paragraph{Baseline information.}
We report the accuracy of KCL and TSC on different benchmark datasets from \cite{li2021targeted}.
For SwAV\footnote{SwAV offical implementation: \url{https://github.com/facebookresearch/swav}.}~\cite{caron2020unsupervised}, PCL\footnote{PCL offical implementation: \url{https://github.com/salesforce/PCL}.}~\cite{PCL} and BYOL\footnote{BYOL offical implementation: \url{https://github.com/deepmind/deepmind-research/tree/master/byol}.}~\cite{grill2020bootstrap}, we use their official open-source implementations.

\end{document}